\pdfoutput=1 

\documentclass[
     a4paper,
     twocolumn
     ]{article}

\usepackage{IEK10} 
\usepackage{natbib}

\usepackage[export]{adjustbox}

\newcommand{\mli}[1]{\mathit{#1}}
\newcommand{\ronos}[1]{\text{RON\,$+$\,OS}}
\newcommand{\untereinander}[2]{\begin{tabular}{c}#1\\#2\end{tabular}}
\newcommand{\untereinanderstretch}[2]{\begin{tabular}{c}#1\vspace{-0.4cm}\\#2\end{tabular}}

\newcommand{\mytitle}{Graph Machine Learning for Design of High-Octane Fuels}

\newcommand{\affil}{
  \begin{itemize}[leftmargin=3mm, itemsep=0mm]
        \item[$^a$]RWTH Aachen University, Process Systems Engineering (AVT.SVT), Aachen 52074, Germany %
		\item[$^b$]Aarhus University, Department of Computer Science, Aarhus, Denmark %
		\item[$^c$]Delft University of Technology, Department of Chemical Engineering, Delft 2629 HZ, The Netherlands %
		\item[$^d$]RWTH Aachen University, Chair of Computer Science 7, Aachen 52074, Germany %
		\item[$^e$]Delft University of Technology, Delft Bioinformatics Lab, Intelligent Systems, 2628 XE Delft, The Netherlands %
		\item[$^f$]RWTH Aachen University, Chair of High Pressure Gas Dynamics, Aachen 52074, Germany %
		\item[$^g$]Forschungszentrum J\"ulich GmbH, Institute for Energy and Climate Research IEK-10: Energy Systems Engineering, J\"ulich 52425, Germany%
		\item[$^h$]JARA-ENERGY, Aachen 52056, Germany
  \end{itemize}
}

\def\firstAuthor{Jan Rittig, Martin Ritzert}
\newcommand{\myauthor}{
	Jan G. Rittig$^{a*}$, 
	Martin Ritzert$^{b*}$, 
	Artur M. Schweidtmann$^c$, %
	Stefanie Winkler$^d$, %
	Jana M. Weber$^e$,  %
	Philipp Morsch$^f$, %
	K. Alexander Heufer$^f$,  %
	Martin Grohe$^d$, %
	Alexander Mitsos$^{h,a,g}$, %
	Manuel Dahmen$^{g\dagger}$ %
}
\newcommand{\correspondingauthor}{
    Corresponding author, E-mail: m.dahmen@fz-juelich.de
}
\author{\myauthor}

\usepackage[hyphens]{url}
\usepackage[
  colorlinks,
  linkcolor=blue,
  citecolor=blue,
  urlcolor=blue,
  pdftitle={\mytitle},
  pdfauthor={\firstAuthor}
]{hyperref}
\usepackage[capitalise, nameinlink]{cleveref}
\crefname{table}{Tab.}{Tab.}

\begin{document}

\twocolumn[
\begin{@twocolumnfalse}
	\thispagestyle{firststyle}
	
	\begin{center}
		\begin{large}
			\textbf{\mytitle}
		\end{large} \\
		\vspace{0.2cm}
		\myauthor
	\end{center}
	
	\vspace{-0.2cm}
	
	\begin{footnotesize}
		\affil
	\end{footnotesize}
	
	\vspace{0cm}

	\section*{Abstract}
	
	Fuels with high-knock resistance enable modern spark-ignition engines to achieve high efficiency and thus low CO$_2$ emissions.
	Identification of molecules with desired autoignition properties indicated by a high research octane number and a high octane sensitivity is therefore of great practical relevance and can be supported by computer-aided molecular design (CAMD).
	Recent developments in the field of graph machine learning (graph-ML) provide novel, promising tools for CAMD. 
	We propose a modular graph-ML CAMD framework that integrates generative graph-ML models with graph neural networks and optimization, enabling the design of molecules with desired ignition properties in a continuous molecular space. 
	In particular, we explore the potential of Bayesian optimization and genetic algorithms in combination with generative graph-ML models.
	The graph-ML CAMD framework successfully identifies well-established high-octane components.
	It also suggests new candidates, one of which we experimentally investigate and use to illustrate the need for further auto-ignition training data.

	\vspace{1.3cm}
\end{@twocolumnfalse}
]


\section{Introduction}

With a share of 23\% of total CO$_2$ emissions, transportation is a major CO$_2$ emission source~\citep{IEA.2020}.
Replacing fossil fuels with renewable alternatives may provide a path towards carbon neutrality for the transportation sector and is investigated actively~\citep{Dahmen.2016, Leitner.2017, Gschwend.2019, Konig.2021}.
An important step towards renewable fuels is the search for suitable gasoline substitutes for use in advanced high compression, turbocharged \emph{spark-ignition}~(SI) engines.
A property of paramount importance for a renewable SI engine fuel is knock resistance, traditionally indicated by the \emph{research octane number}~(RON)~\citep{AmericanSocietyforTesting.2018}, the \emph{motor octane number}~(MON)~\citep{AmericanSocietyforTesting.2019}, and more recently the \emph{octane sensitivity}~(OS), i.e., the difference between RON and MON values. 
The weighted sum of RON and OS is referred to as the \emph{octane index} (OI)~\citep{Kalghatgi.2001a}.
For modern SI engines, fuels with both high RON and high OS, hence high OI, are desired as they enable engine operation at conditions associated with particularly high efficiency~\citep{Kalghatgi.2005, Bell.2010, Mittal.2008, Amer.2012, Kalghatgi.2014, Szybist.2017, AbdulManan.2018}.
To boost the OI of a fuel, chemical species with high RON and high OS such as ethanol and MTBE can be added~\citep{Demirbas.2015, Badia.2021}.
Identification of further molecules providing octane boosting is of great practical relevance and is studied actively, e.g., see~\citep{Badia.2021, Li.2022}.
Herein, we aim to identify such promising candidates exhibiting both high RON and high OS by \emph{computer-aided molecular design} (CAMD).
In particular, we investigate the role of novel methods from the domain of \emph{graph machine learning} (graph-ML).

Traditionally, the search for molecules with desired properties for a given application has been mostly guided by human experts and experimentation.
CAMD can enhance this process by utilizing computational methods to efficiently pre-screen a large number of molecular structures so that experiments can be dedicated to the most promising candidates. 
A wide variety of methods and tools for CAMD has been proposed over the last decades; we refer the interested reader to review articles for a detailed CAMD overview~\citep{joback1989designing, Achenie.2003, Gani.2004, Zhang.2015, Ng.2015, Austin.2016, Alshehri.2020}.
Generally, the CAMD process incorporates the computational generation of candidate structures and the model-based prediction of their physico-chemical properties.
Well-established approaches for the generation of candidate structures include formulating optimization problems in which structural groups are pieced together to form molecules~\citep{Samudra.2013, Austin.2016}, exhaustive generation of molecular structures in a sequential generate-and-test manner~\citep{Harper.1999}, and utilizing evolutionary theory to evolve molecular structures~\citep{Douguet.2005}. 
For predicting application-relevant properties of the formed candidate structures, CAMD typically employs quantitative structure-property relationships~(QSPRs)~\citep{Katritzky.2010}. 
QSPRs first describe the molecular structure by so-called molecular descriptors, e.g., atom counts, and secondly map those descriptors to a property of interest by linear or nonlinear models.
Today, nonlinear ML models such as feedforward neural networks or random forests are often utilized in this regression step~\citep{Mitchell.2014, Lo.2018, Muratov.2020}.

For classical CAMD, a broad range of applications~\citep{Alshehri.2020} can be found in the process systems engineering (PSE) literature, covering the design of single molecules (e.g., ionic liquids~\citep{Karunanithi.2013}, polymers~\citep{Zhang.2015}), the design of mixtures~\citep{Austin.2017, Austin.2018, Liu.2019}, as well as integrated product and process design~\citep{Lampe.2015, Schilling.2017}.
Classical CAMD techniques have also been applied extensively in the context of fuel design~\citep{Dahmen.2016, Hechinger.2010, Hoppe.2016, Whitmore.2016, McCormick.2017, Lunderman.2018}.
For example, in two previous articles~\citep{Dahmen.2012, Hoppe.2016}, we used enumeration-based generation of oxygenated hydrocarbons and subsequently screened the obtained molecules via QSPR models with respect to engine-relevant properties. 
We previously also developed a generate-and-test approach where molecular candidates are generated by iteratively refunctionalizing bioderived intermediates based on pre-defined transformation rules~\citep{Dahmen.2016}.
Also, Cai et al.~\citep{Cai.2021} proposed a gasoline design model that employs rule-based transformation of molecules in combination with QSPR for property prediction to identify molecules with desired fuel properties such as high RON. 

ML has recently been utilized for molecular structure generation by means of generative ML models, leading to novel, fully \emph{ML-based CAMD} approaches~\citep{Alshehri.2020, Elton2019}.
In generative ML for molecules, two main directions can be distinguished: String-based approaches, e.g., based on SMILES strings~\citep{Weininger.1988}, and graph-based approaches, the latter directly working on the molecular graph.
For both directions, a range of models has been developed such as recurrent neural networks~(RNNs), variational or adversarial autoencoders~(VAEs/AAEs), generative adversarial networks~(GANs), and reinforcement learning~(RL)~\citep{Elton2019, Faez.2021}.
The goal of such generative ML techniques is the unsupervised learning from a data set of molecular structures to generate new, chemically feasible structures that were not seen during training, thereby designing molecules.
Specifically, generative ML models typically learn to encode molecules into a continuous space, the so-called latent space, and then decode samples from the latent space back to molecular structures. 
The continuous latent space is assumed to capture chemical information about molecules and embed molecules with similar structure or even similar properties close to each other~\citep{Winter.2019a}.
Depending on the model architecture, ML-based CAMD typically relies either on strategic sampling of molecules from the latent space of the generative model using optimization strategies, e.g., with VAEs~\citep{Sanchez-Lengeling2018, Jin2018, Kajino2019}, or on direct generation of molecules with desired properties, e.g., by GANs~\citep{Guimaraes2018, DeCao2018} or RL~\citep{You2018, Zhou2019}.
In contrast to classical CAMD, generative models in ML-based CAMD replace discrete molecule representations such as combinations of structural groups, molecular graphs, or SMILES strings with a continuous representation, thus enabling the use of continuous optimization approaches for molecular design~\citep{Coley.2020a}.

ML has also recently enabled end-to-end learning of physico-chemical properties from molecular structure by means of graph neural networks (GNNs)~\citep{Sperduti1997, Gori2005, Scarselli2009}.
GNNs are graph-ML architectures that directly operate on the underlying graph structure of a molecule and thus circumvent the need for selecting meaningful molecular descriptors, a step that is inherent to all QSPR/QSAR approaches.
Instead, GNNs enable a data-driven end-to-end learning framework for molecular property prediction.

Up to now, fully ML-driven CAMD has mainly focused on drug design~\citep{Elton2019, Xia.2019, Xiong.2021, Gaudelet.2021}.
A particular reason might be the availability of large training data sets and the incorporation of multiple drug design targets such as logP and drug-likeness in benchmarking platforms such as MOSES~\citep{Polykovskiy.2020} and GuacaMol~\citep{Brown.2019}.
Such ML-driven CAMD approaches often combine molecule generation and property prediction (e.g., VAEs~\citep{Jin2018, Kajino2019}), and sometimes optimization (e.g., GANs~\citep{Guimaraes2018, DeCao2018} or RL~\citep{Zhou2019}) in a single ML model which needs to be retrained once the design target property changes and typically requires large property data sets for training.

In contrast to drug design, PSE applications, in particular model-based fuel design, often take place in a data-scarce environment, making ML-based CAMD challenging.
In fact, there is only one very recent study using generative ML for fuel design:
Liu et al.~\citep{Liu.2022} employed a string-based VAE to generate a large database of non-oxygenated hydrocarbons for subsequent screening of candidates with respect to fuel properties, followed by sampling further candidates from the most promising regions of the VAE's latent space.
However, ML-driven CAMD has not yet been utilized for fuel design focusing on high SI engine efficiency including oxygenated hydrocarbons.
Moreover, graph-ML approaches have not yet been applied to computer-aided fuel design.

\begin{figure*}
	\centering
	\includegraphics[width=1\textwidth, trim={0cm 3cm 0cm 2.9cm},clip]{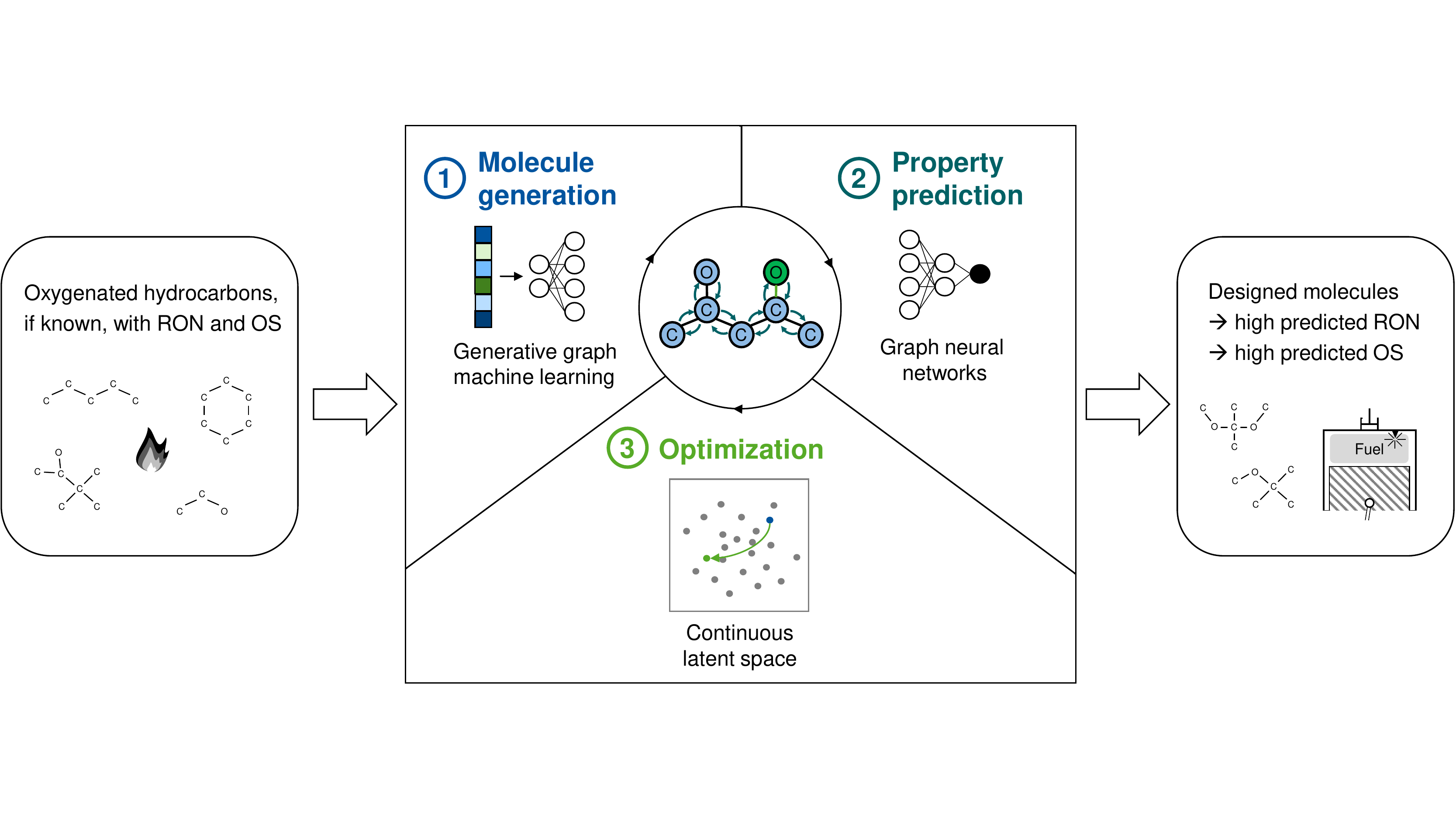}
	\caption{Schematic overview of the modular graph-ML CAMD framework for identification of high-octane fuels.}
	\label{fig:DesignLoopSimple}
\end{figure*}

In the present contribution, we propose a modular graph-ML CAMD framework\footnote{Code is openly available, see~\citep{Graph_ML_Fuel_Design_GIT} \label{CODE-URL}} that integrates state-of-the-art graph-based ML methods and tools from the ML and drug design community and apply our framework to computer-aided design of high-octane fuel components for SI engines.
Our framework is depicted in Figure~\ref{fig:DesignLoopSimple} and consists of three distinct modules:
(1)~molecule generation by generative graph-ML models that learn a continuous molecular space from which new molecules can be generated;
(2)~property prediction through our recently published GNN model for fuel ignition quality prediction~\citep{Schweidtmann2020_GNNs};
(3)~optimization for strategic sampling from the continuous space of the generative graph-ML models to identify vectors that correspond to molecules with high predicted RON and OS values.
Our framework has a modular architecture requiring minimal changes to the model structures if an additional property shall be targeted, i.e., only a new property model needs to be trained and added, but the molecule generation and optimization modules do not need to be altered.
Thus, the modular setup enhances reusability and therefore reduces the training effort compared to a single ML model approach, as indicated by Winter et al.~\citep{Winter.2019b}.

We explore three different generative graph-ML models and two different optimization strategies. 
Importantly, we propose an applicability domain approach for GNN-based property prediction that allows us to focus the design process on molecules that presumably come with reliable predictions.
We analyze the influence of the different ML methods on the structure and properties of the resulting molecules and compile a list of most promising high-octane fuel candidates.
Finally, we perform an experimental investigation of one selected high-octane fuel candidate that emphasizes the importance of experimental validation of CAMD results and discuss potential pitfalls of the fully data-driven approach, particularly in a data-scarce environment.

The article is structured as follows: 
In the section ``Preliminaries of graph machine learning'', we briefly introduce the main principles behind graph-ML for molecules with regard to both molecule generation and property prediction.
In the subsequent section, we present the modular graph-ML CAMD framework for design of high-octane fuels.
The application of the framework in the section ``Results and discussion'' includes a comparative analysis of the candidates obtained with different graph-ML modules and the experimental investigation of one particular candidate.
The last section concludes our work.

\section{Preliminaries of graph machine learning}\label{sec:Fundamentals}
Graph-ML relies on a graph representation of molecules that can be utilized for generating molecular structures from a continuous space and for property prediction, as we briefly describe in the following.
The interested reader is referred to~\citep{Hamilton2017, Wu2020, Bronstein.2021} for further details on graph-ML.

\subsection{Molecular graph}
The \emph{molecular graph} of a molecule is an undirected graph $G_\text{mol} = \{V, F_v, E, F_e\}$; the nodes $V$ represent the atoms; pairs of atoms $u,v\in V$ that share a bond are connected by edges $(u,v)\in E$.
Additional features of nodes (e.g., type of atom, degree of hybridization) are stored in $F_v$, while additional features of edges (e.g., bond length or type) are stored in $F_e$.

\subsection{Generative models}\label{subsec:MolGen}

Generative ML, the unsupervised learning from input data to generate new data that is similar to the provided data, allows to perform fully data-driven molecule generation and is an active research area~\citep{Elton2019, Faez.2021, Gaudelet.2021, Atz.2021}.
Various works have developed string-based ML models in order to generate molecules with optimal properties based on SMILES~\citep{Kadurin2017, Gomez-Bombarelli2018, Krenn2020, Blaschke2018, Lim2018, Bjerrum2018, Prykhodko2019, Griffiths2020}, InChI~\citep{Winter.2019a}, or SELFIES~\citep{Krenn2020}, the latter being a more robust string representation of molecules.
In contrast, graph-ML directly works on the molecular graph which is arguably the more natural representation of a molecule and provides permutation invariance~\citep{Mercado.2021}, i.e., there is exactly one molecular graph for each molecule (neglecting steric effects).
In this paper, we focus on two frequently employed generative graph-ML approaches~\citep{Elton2019, Faez.2021, Gaudelet.2021}: VAEs and GANs.
Both methods construct a latent space where molecules are encoded as high-dimensional continuous vectors, referred to as latent vectors (LVs), which we denote as $\mathbf{h}_\text{LV} \in \mathbb{R}^n$ with the dimension $n$ being a hyperparameter.
We denote the encoding of a molecular graph into the latent space as a function
\begin{equation}
e_\text{GEN} \colon G_\text{mol} \mapsto \mathbf{h}_\text{LV}.\label{eq:genEnc}
\end{equation}
\begin{figure}
	\begin{subfigure}[c]{0.45\textwidth}
		\centering
		\includegraphics[width=\textwidth, trim={7.4cm 5.8cm 8.6cm 5cm},clip]{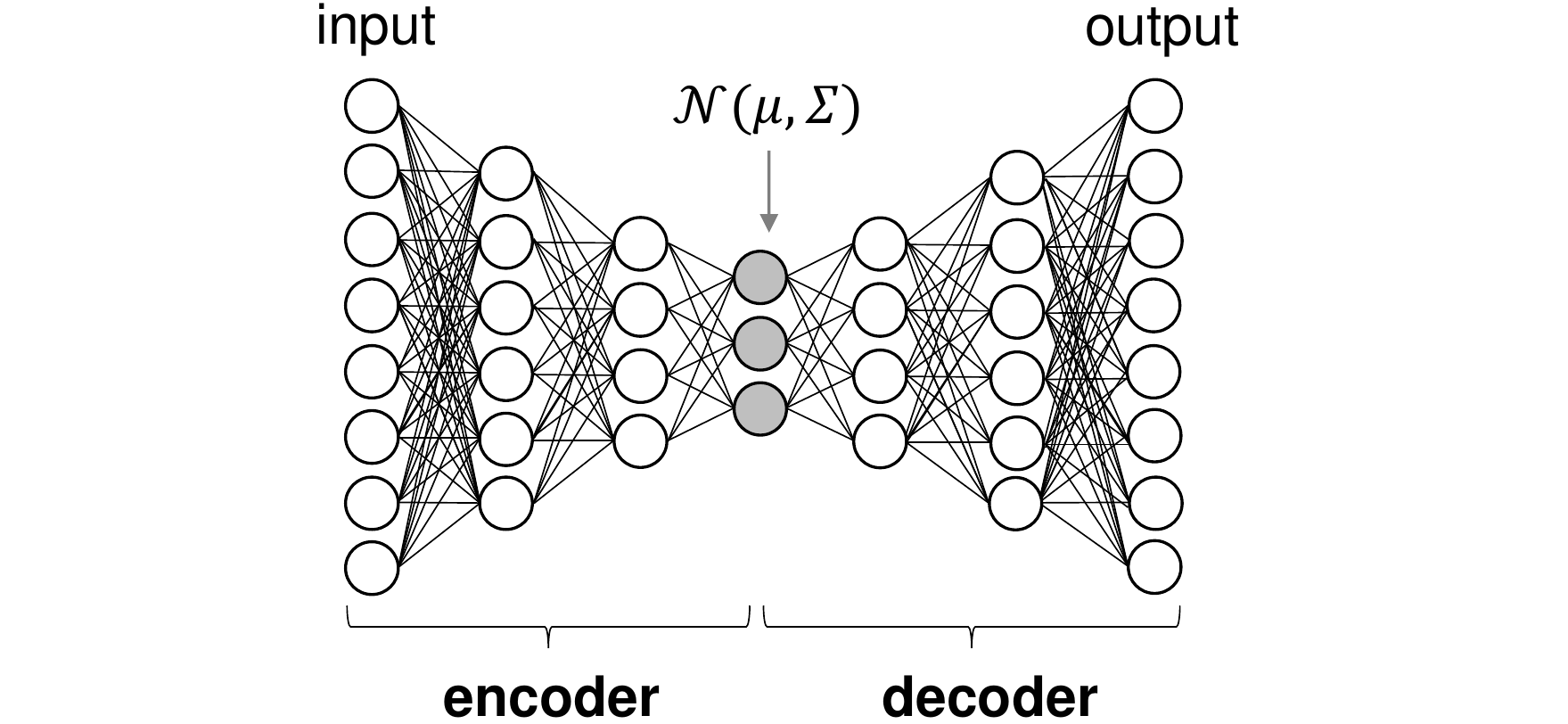}
		\subcaption{VAE}
	\end{subfigure}
	\hfill
	\begin{subfigure}[c]{0.45\textwidth}
		\centering
		\includegraphics[width=\textwidth, trim={8cm 6cm 8cm 4cm},clip]{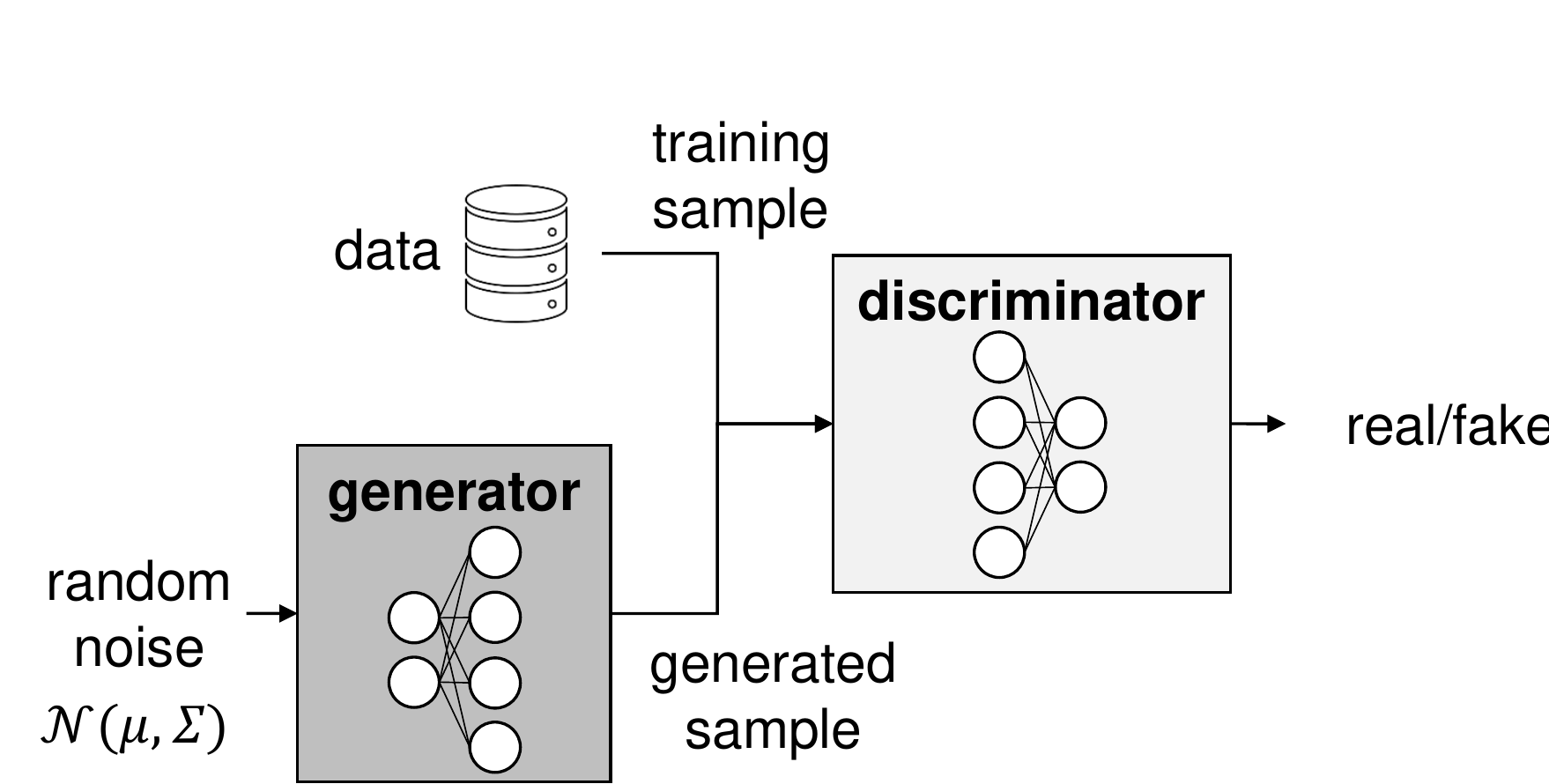}
		\subcaption{GAN}
	\end{subfigure}
	\caption{Schematic structure of (a) VAEs and (b) GANs.}
	\label{fig:GeneratorStructure}
\end{figure}
Autoencoders, and specifically VAEs, are a class of neural network architectures that employs an hourglass shape (cf. Figure~\ref{fig:GeneratorStructure}a).
They are trained to reproduce the input data at the output layer, a non-trivial task as the information has to be moved through some narrow layers in the middle of the network, that is, the hourglass shape forces VAEs to learn $\mathbf{h}_\text{LV}$ as a low-dimensional representation of the input data at the most narrow layer.
The left part of the network (from input to the latent vector) is called the \emph{encoder} and the right part (from the latent vector to the output) is referred to as the \emph{decoder}.
The main difference between a standard autoencoder and a variational autoencoder (VAE) is that the latter assumes an underlying distribution for the data that it tries to learn in the latent vector space, e.g., a multivariate Gaussian distribution $\mathbf{h}_\text{LV} \sim \mathcal{N}(\mu,\,\Sigma)\,$ with parameters $\mu$ and $\Sigma$.
VAEs can therefore be used to generate new data from presumably the same distribution as the input data.
In the molecular context, VAEs map discrete molecule representations such as graphs to a continuous distribution from which new molecules can be sampled.

GANs generate objects from a latent representation in a different manner (cf. Figure~\ref{fig:GeneratorStructure}b). 
Instead of trying to reproduce an input sample, a GAN consists of two neural networks, a \emph{generator} and a \emph{discriminator}, where the discriminator is trained to distinguish between output data produced by the generator and real data, i.e., the training samples.
The generator thus learns to produce output data that resembles a given training data based on random input vectors $\mathbf{h}_\text{LV}$ that are, for example, sampled from a Gaussian distribution, i.e., $\mathbf{h}_\text{LV} \sim \mathcal{N}(\mu,\,\Sigma)\,$.
In a GAN, the latent space therefore corresponds to the input space of the generator.
We denote the decoding of the latent vector $\mathbf{h}_\text{LV}$ to the molecular graph in case of both generators, VAE and GAN, with the function
\begin{equation}
d_\text{GEN} \colon \mathbf{h}_\text{LV} \mapsto G_\text{mol}.\label{eq:genDec}
\end{equation}
%

\subsection{Graph-based property prediction}\label{subsec:ProPre}
A GNN~\citep{Gori2005, Scarselli2009} is a type of neural network that operates directly on the graph structure and thus enables end-to-end learning in molecular property prediction.
Thereby, GNNs avoid the need for the often subjective manual selection process of molecular descriptors in QSPR/QSAR modeling that requires intuition and experience of the modeler. 
\begin{figure}
	\centering
	\includegraphics[width=0.5\textwidth, trim={2cm 4cm 2cm 5cm}, clip]{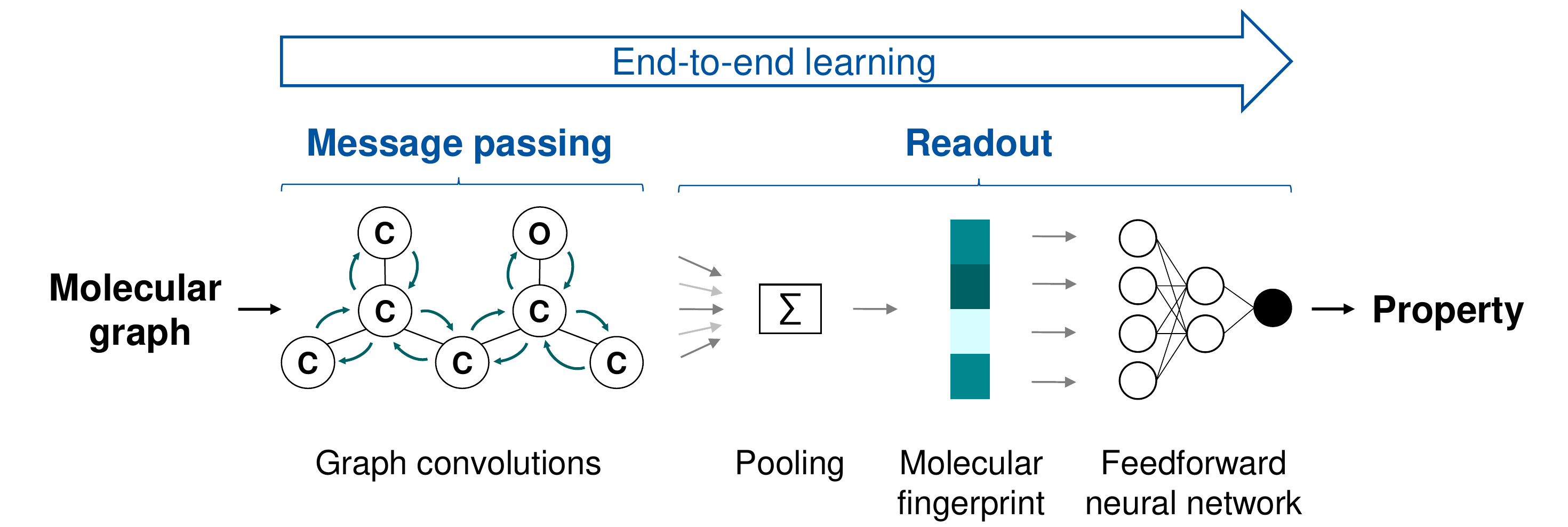}
	\caption{Schematic structure of a graph neural network for molecular property prediction.}
	\label{fig:GNNstructure}
\end{figure}

GNNs for molecular property prediction are typically structured into two parts, a message passing phase and a readout phase~\citep{Gilmer2017, Coley2017} (cf. Figure~\ref{fig:GNNstructure}). 
In the message passing phase, structural information is extracted from a local neighborhood of atoms by means of graph convolutions.
In each graph convolution, every node sends a message to all its neighbors and thus also receives a message from each of its neighbors.
The node uses the received messages, typically in form of a weighted sum, to update its current state (e.g., in GCN \citep{Hamilton2017} and GAT \citep{VelickovicCCRLB18}).
The update of the state $\mathbf{h}_v^l$ of a node $v$ in a graph convolutional layer $l$ can then be written as 
\begin{equation}
\mathbf{h}_v^{l+1} = \sigma_\text{ReLU}\left( \mathbf{h}_v^l \mathbf{W}_1  + \sum_{u\in N(v)} \mathbf{h}_u^l \mathbf{W}_2\right)
\label{eq:GNN_UptMes},
\end{equation}
\noindent where $\mathbf{W}_1,\mathbf{W}_2$ are trainable weight matrices, $N(v)$ is the one-hop neighborhood of $v$, and $\sigma_\text{ReLU}$ denotes the elementwise application of the ReLU activation function.
Many different update functions have been proposed in the last years, see, e.g.,~\citep{Wu2020, Zhou2020, Zhang2020_Rev}, to advance the basic Equation~\eqref{eq:GNN_UptMes} into a more powerful model for extracting information from the graph during message passing~\citep{Xu.2018}.
For instance, inter-atomic distances and angles between atom pairs \citep{Schutt2018, Unke.2019, Klicpera.06.03.2020, Zhang2020_MXM} are commonly considered.
Higher-order GNNs~\citep{Morris2019, Flam-Shepherd2020} and approaches where the information exchange is also based on individual edges~\citep{Yang2019} constitute further extensions to the basic GNN approach.

Subsequent to the message passing phase, a GNN employs a readout phase, where the molecular structure information that is stored in the nodes is aggregated into a single vector for the complete molecule, the so-called \emph{molecular fingerprint} $\mathbf{h}_\text{FP}$.
This aggregation, also called pooling, is typically performed by summing up the states of all nodes in the molecular graph after the last graph convolutional layer $L$, i.e., \(\mathbf{h}_\text{FP}~=~\sum_{v \in V}~\mathbf{h}_v^L \).
We denote the GNN encoding of the molecular graph into the molecular fingerprint with the function
\begin{equation}
g_\text{GNN} \colon G_\text{mol} \mapsto \mathbf{h}_\text{FP} \label{eq:GNN_Enc}.
\end{equation}
Note that although the molecular fingerprint $\mathbf{h}_\text{FP}$ in a GNN and the latent vector $\mathbf{h}_\text{LV}$ in a generative ML model both represent a molecule in a continuous space, they are not related.
In the GNN, the molecular fingerprint $\mathbf{h}_\text{FP}$ is passed through a feedforward neural network (cf. Figure~\ref{fig:GNNstructure}) to yield the property prediction \( \hat{p}~=~\mli{MLP}(\mathbf{h}_\text{FP}) \).
Here, a multi-layer perceptron (MLP) is one of the simplest feedforward neural architectures and most frequently employed.
We denote the entire end-to-end prediction process of a GNN as a function $f_\text{GNN}$ that maps the molecular graphs to a property prediction, i.e.,
\begin{equation}
f_\text{GNN} \colon G_\text{mol} \mapsto \hat{p}\label{eq:GNN_FFP}.
\end{equation}

\section{Graph-ML CAMD framework for high-octane fuels}\label{sec:Modeling}

\begin{figure*}
	\centering
	\includegraphics[width=\textwidth, trim={0cm 3cm 0cm 1.5cm},clip]{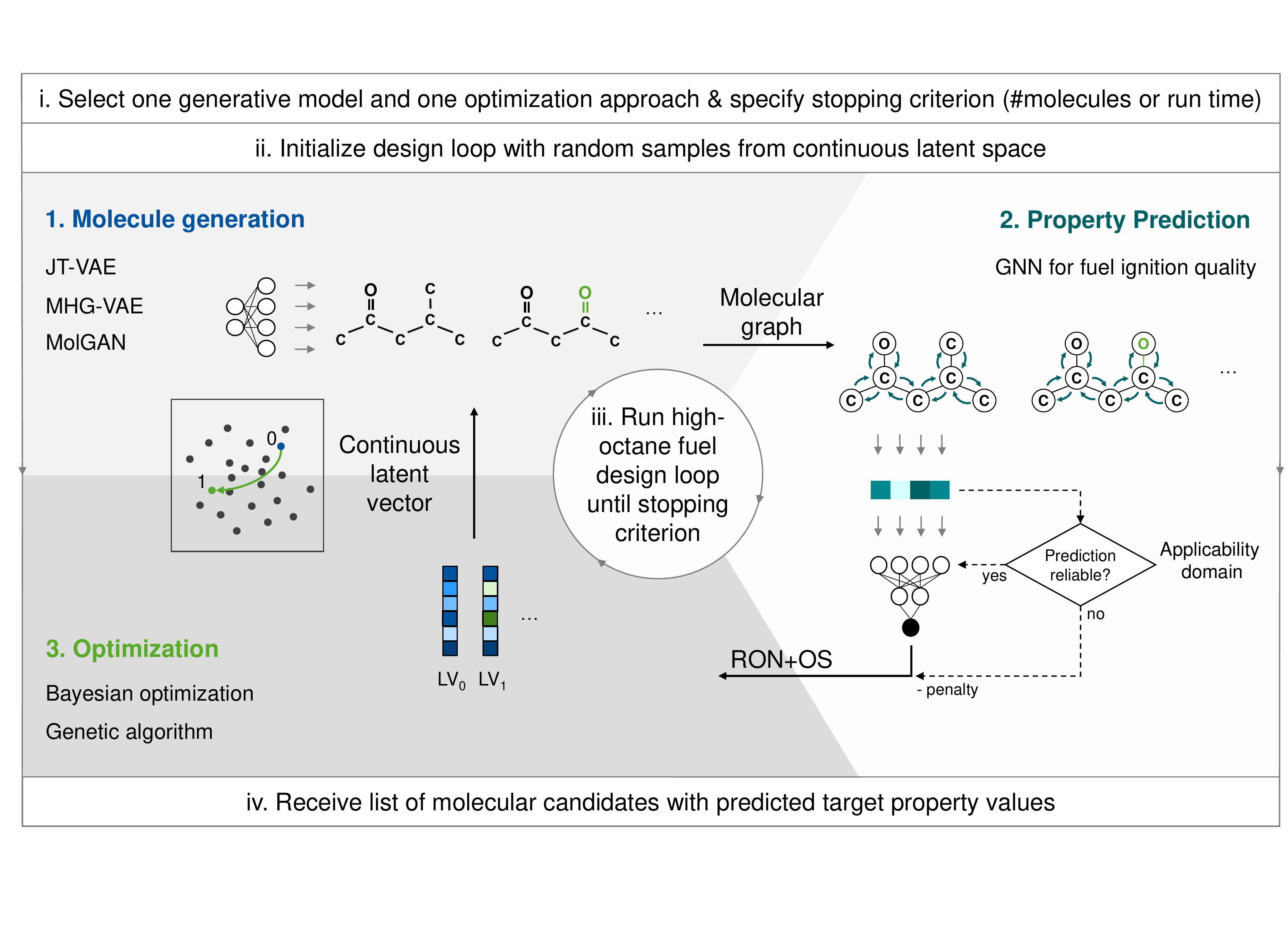}
	\caption{Detailed overview of the modular graph-ML CAMD framework for identification of high-octane fuels including methods for the individual modules.}
	\label{fig:DesignLoop}
\end{figure*}

In this section, we propose a fully data-driven, modular graph-ML CAMD framework for identification of high-octane fuels.
The framework utilizes recent methods from the field of generative graph-ML and GNNs to design molecules with high-knock resistance for modern SI engines.
Specifically, we set out to maximize the sum of RON and OS, hence the OI, as high-efficiency SI engines require both a high RON and a high OS~\citep{Kalghatgi.2005, Bell.2010, Mittal.2008, Amer.2012, Kalghatgi.2014, Szybist.2017, AbdulManan.2018}. 
We show a high-level overview of our framework in Figure~\ref{fig:DesignLoopSimple} and provide a detailed framework overview including our choices for algorithms and models in the three different modules in Figure~\ref{fig:DesignLoop}.
We combine the three modules to form an iterative molecular design loop:
The \emph{optimization module} proposes initial latent vectors from a continuous space, $\mathbf{h}_\text{LV}$, that are translated to corresponding molecules by the \emph{molecule generation module}, cf. Equation~\eqref{eq:genDec}. 
Then, the \emph{property prediction module} performs the property evaluation, cf. Equation~\eqref{eq:GNN_FFP}, and based on the property predictions, the optimization algorithm suggests new latent vectors to be tested. 
This iterative procedure is repeated until a pre-defined stopping criterion is met, e.g., a certain number of molecules has been evaluated.

An important observation with the graph-ML CAMD framework though is that not all molecules come with physically reasonable predictions.
For instance, we have observed a molecule with predicted OS $> 400$ and negative RON and negative MON\footnote{Found by GA when optimizing for OS only}.
In fact, the optimization often exploits weak spots of the GNN prediction model.
Those weak spots typically appear for molecules that are strongly dissimilar from the molecules used for training the GNN.
To focus on molecules with more reasonable property predictions, we extend the iterative design loop by an \emph{applicability domain} (AD) for the GNN property prediction model.
To this end, we build upon the AD approach from our previous study~\citep{Schweidtmann2021_AD} where we proposed to use a one-class classification model to identify the AD of feedforward NNs.
The classification model learns from the data on which the NN is trained to decide if a new data point is similar to the training data and thus considered within the input domain for which the NN presumably provides reliable predictions.
To transfer the AD approach to GNNs, we apply the classification model to the molecular fingerprint that serves as input to the MLP part of the GNN (cf. Subsection ``Graph-based property prediction''). 
If the AD is included, GNN predictions considered unreliable by the AD are ignored and instead a penalty value (-1000) is returned to the optimization approach so that the corresponding molecules are assigned a low objective value.

The design loop runs can be formulated as an optimization problem that aims to find the molecules with the highest predicted value of a certain target property $\hat{p}$ of interest, i.e.,
\begin{equation}
\begin{aligned}
&\!\max_{\mathbf{h}_\text{LV}}   &\qquad     & \hat{p}\\
&\text{s. t.}       &           & G_\text{mol} = d_\text{GEN}(\mathbf{h}_\text{LV}),\\
&                   &           & \hat{p} = f_\text{GNN}(G_\text{mol}),\\
&                   &           & \mathbf{h}_\text{FP} = g_\text{GNN}(G_\text{mol}),\\
&                   &           & \mli{AD}(\mathbf{h}_\text{FP}) \geq 0,
\end{aligned}\label{eq:RONOS_constrOpt}
\end{equation}
whereby the constraint with $\mli{AD}(\mathbf{h}_\text{FP}) \geq 0$ denotes a positive decision by the AD model.

Due to the high dimensionality of the search space that corresponds to the latent space of the generator models (see Equation~\eqref{eq:RONOS_constrOpt}), deterministic global optimization is too computationally costly and practically impossible with current methods (cf. Subsection ``Optimization'' below).
Instead, we employ black-box optimization approaches that direct a heuristic search towards molecules with high $\hat{p}$. 
Note that uncertainties in the prediction model prohibit a strict ranking of molecular candidates with similar $\hat{p}$ values.
Practically, we therefore compile a list of molecules sampled by the optimizer and perform an investigation of the top candidates, i.e., the molecules with the highest $\hat{p}$ values.
Having multiple top candidates, also allows to take additional desired properties into account in later investigations, e.g., availability for procurement and low production costs.

In the following, we briefly describe the three generative graph-ML models used in this paper for the generation module, the GNN model used for the property prediction module, the two optimization algorithms used in the optimization module, and our AD approach.

\subsection{Molecule generation}\label{subseq:ModMolGen}
We consider two graph VAE models as generators: The Junction-Tree VAE by Jin et al.~\citep{Jin2018} (JT-VAE) and the Molecular Hypergraph Grammar VAE by Kajino~\citep{Kajino2019} (MHG-VAE).
Furthermore, we employ MolGAN, a GAN for molecular graphs published by De Cao and Kipf~\citep{DeCao2018}.
Those three models have close to 100\% chemical validity, i.e., almost 100\% of the generated molecules are chemically feasible~\citep{Jin2018, Kajino2019, DeCao2018}, a feature that earlier generative methods struggled with, cf.~\citep{Kusner.2017, Dai.2018}.
Apart from achieving high validity, the three models have strong conceptual differences, presumably leading to molecules with somewhat different characteristics.

The JT-VAE~\citep{Jin2018} utilizes two graph representations of a molecule in parallel: The molecular graph and its associated junction tree, which is a contracted cycle-free graph generated by merging cycles of atoms into a single node. 
For encoding, the JT-VAE learns molecular structure information, represented as high-dimensional vectors, from the molecular graph and the junction tree through graph convolutions (cf. Section ``Preliminaries of graph machine learning'').
For decoding, first, the junction tree's latent vector is decoded resulting in the general molecular structure. 
Then, the molecular graph's latent vector is decoded to determine the characteristics of the nodes within the junction tree, i.e., (re)generating the local structure of the molecule.
Jin et al. report a molecule reconstruction rate of 76.7\% and 100\% chemical validity of the decoded molecules~\citep{Jin2018}.

The MHG-VAE~\citep{Kajino2019} generates a graph grammar from the given training molecules which is used for the reconstruction of molecules.
In this automatically generated graph grammar, terminal symbols can refer to either single atoms or complete functional groups and the rules of the grammar describe how such atoms of partial molecules can be combined into a chemically valid molecule.
During the generation of the grammar, MHG-VAE ensures that the grammar accounts for chemical feasibility constraints such as valency rules, explaining the validity of 100\%.

\begin{figure}
	\centering
	\includegraphics[width=0.45\textwidth, trim={8cm 5cm 8cm 5cm},clip]{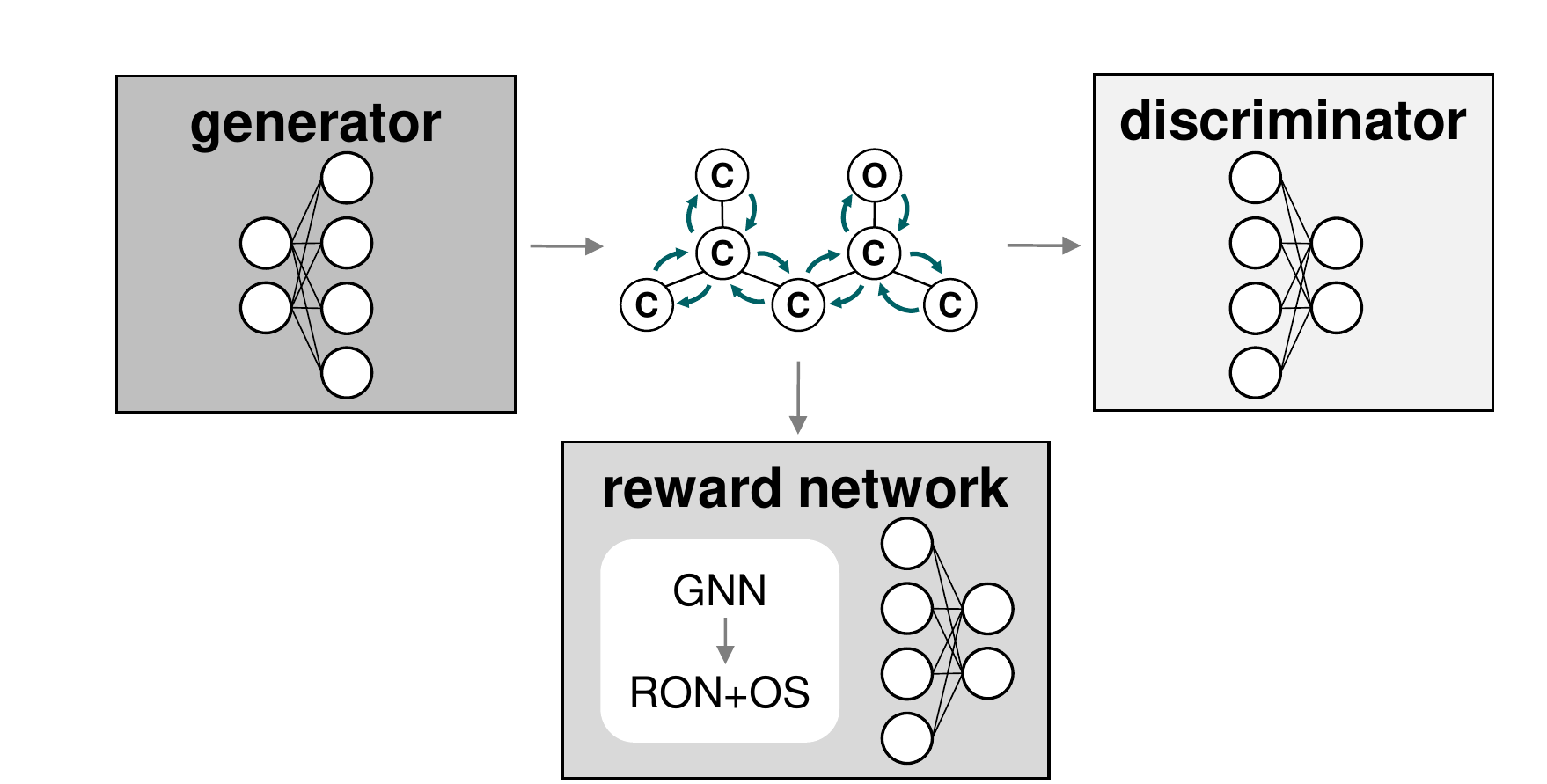}
	\caption{Adapted MolGAN for high-octane fuels, modified from~\citep{DeCao2018}. The reward network is coupled with our GNN~\citep{Schweidtmann2020_GNNs} for predicting RON and OS values.}
	\label{fig:FuelGAN}
\end{figure}

MolGAN~\citep{DeCao2018} only partially relies on graphs.
Its adaptation to our case of high-octane fuel design is illustrated in Fig.~\ref{fig:FuelGAN}.
The generator tries to directly predict a molecular graph's adjacency matrix with corresponding atom and bond features by using an MLP with a fixed output size, i.e., the maximal size of a molecule that can be predicted by MolGAN is bounded.
On the other hand, the discriminator is a GNN.
One conceptual difference to the VAEs is that MolGAN is able to focus the generation on molecules with desirable properties by using a `reward network', i.e., a third network that encourages the generator to output molecules with high RON and OS.
We use our GNN model~\citep{Schweidtmann2020_GNNs} to provide RON and OS predictions such that, in contrast to the VAEs, the training of MolGAN partially depends on the property prediction module.
De Cao \& Kipf state that while MolGAN generates novel molecules with desirable properties and almost 100\% chemical validity, it also outputs many duplicates with only about one in ten molecules being unique~\citep{DeCao2018}.

\subsection{Property prediction}\label{subseq:ModPropPred}
We recently developed a GNN for predicting the RON, MON, and the derived cetane number (DCN) of a wide range of oxygenated and non-oxygenated hydrocarbons~\citep{Schweidtmann2020_GNNs}, e.g., (cyclo-) alkanes, (cyclo-) alkenes, alcohols, esters, ethers, aromatics, and ketones.
The model architecture is based on higher-order GNNs~\citep{Morris2019} and additionally leverages the increased stability and accuracy of ensemble methods~\citep{Breiman.1996, Breiman.1996b}, i.e., the final property prediction is the average of multiple higher-order GNN predictions.
Further, our GNN incorporates multi-task learning~\citep{Caruana.1997, Ruder.2017} as it was trained on RON, MON, and DCN values simultaneously allowing the model to capture and exploit correlations between octane and cetane numbers. 

As described in detail in~\citep{Schweidtmann2020_GNNs}, we compiled a data set comprising 335 RON, 318 MON, and 236 DCN values for 505 unique molecules in total to train the GNN. 
85\% of the data was used for training and validation, and 15\% was used for testing.
Note that for most molecules, both RON and MON values and thus OS were available. 
The mean absolute prediction error of the GNN model was 4.5 on the RON test set and 4.4 on the MON test set, indicating an overall high prediction quality on par with state-of-the-art QSPR- and ML-based RON and MON prediction models, cf.~\citep{Schweidtmann2020_GNNs}.
The test sets also contain few outliers: Six predictions for RON and seven predictions for MON have a deviation $>$ 10, which we attribute, similarly to vom Lehn et al~\citep{VomLehn.2020}, to some of these molecules having unique characteristics that are not well represented in the training data, a relatively small number of data points available with low RON and MON values, and potential disruptive factors in experimental data assembled from different sources.

\subsection{Optimization}\label{subseq:ModOpt}

To sample molecules with high RON and OS from the latent space of the generative models, we employ numerical optimization using the \ronos{} score predicted by the GNN model as objective function.
Specifically, we seek to maximize $\hat{p} = \text{RON} + \text{OS} = 2 \cdot \text{RON} - \text{MON}$ (cf. Equation~\eqref{eq:RONOS_constrOpt}).
We explore two derivative-free stochastic global optimization methods to perform the molecule sampling: A Bayesian optimization algorithm and a genetic algorithm.

\emph{Bayesian optimization} (BO) is a probabilistic approach for global optimization~\citep{Shahriari.2016} commonly used for optimization of black-box models that are costly to evaluate.
Usage of BO is well-established in ML-based CAMD, see, e.g.,~\citep{Jin2018, Kajino2019, Gomez-Bombarelli2018}, as well as in chemical engineering applications, e.g., the design of experiments in automated reaction platforms~\citep{Schweidtmann2018_MLflow, Felton.2021, Hase.2021}.
BO uses a surrogate model, typically a Gaussian process (GP), to map the input variables to the objective.
Based on the surrogate model, an acquisition function locates input variable values that have a high potential of maximizing the objective by accounting for both exploitation and exploration.
For running BO, the GP is initialized with a set of feasible points.
Then, the following steps are repeated until a termination criterion is reached: The acquisition function is optimized to determine the next sampling points, the sampling points are evaluated with respect to the objective function, and the objective values are used to refine the surrogate model.
Note that different optimization algorithms can be used for maximizing the acquisition function, cf.~\citep{Shahriari.2016}.

A \emph{genetic algorithm} (GA) is a meta-heuristic, population-based approach for global optimization that is inspired from evolutionary theory~\citep{Holland.1992, Whitley.1994}.
It is typically applied to optimization problems with cheap and fast evaluations of the objective function.
In GAs, a set of feasible points is called population.
Each feasible point has genes corresponding to specific values for the input variables of the optimization problem and constitutes a fitness related to the objective value.  
To solve an optimization problem, an initial population evolves in an iterative manner over multiple generations by promoting points with high fitness and using evolutionary heuristics, e.g., combining genes of high fitness points, to replace points with low fitness.
We choose the fitness to be \ronos{} to directly optimize for high-octane ratings.

A major challenge in ML-based CAMD is the high dimensionality of the generators' latent space which typically requires a large number of sampling points for optimization, e.g., in case of our generative models, we have latent space dimensionalities of 56 (JT-VAE)~\citep{Jin2018}, 72 (MHG-VAE)~\citep{Kajino2019}, and 32 (MolGAN)~\citep{DeCao2018}.
BO, however, employs a GP as surrogate model that in standard form has cubic scaling in complexity with respect to the number of sampling data points.
Following the strategy by Kajino~\citep{Kajino2019}, we thus use PCA to reduce the dimensions of both the JT-VAE and the MHG-VAE before performing BO.
Since the execution time of the evolutionary-based heuristics in the GA does not suffer from a high number of sampling points, we run the GA without dimensionality reduction.
Note that the effects of PCA-based dimensionality reduction on the obtained molecules as well as the use of other mitigation strategies, such as reduction of the latent dimension within the generator or modification of BO for high-dimensional problems, see, e.g.,~\citep{Snoek.2015, Shahriari.2016, Wang.2016, Wang.2018, KirschnerPMLR19}, are beyond the scope of this work.

\subsection{Applicability domain}
The AD of a model is a well-established concept in QSPR/QSAR modeling and is based on the general assumption that the prediction model would provide most reliable predictions for molecules that are similar to the ones seen during training~\citep{Tropsha.2003, Jaworska.2005, Gramatica.2007, Weaver.2008}.
Molecular similarity is usually assessed by means of a distance metric, e.g., the Euclidean distance between the descriptor values of two molecules~\citep{Jaworska.2005, Sheridan.2004}.
For molecular property prediction with GNNs, determination of the AD is largely unexplored. 
Only very recently first approaches to quantify the AD of GNNs based on uncertainty quantification methods were proposed~\citep{Hirschfeld.20.05.2020, Soleimany.2021, Nigam.2021}.
Conceptually, defining the AD of a GNN requires handling the varying input sizes of molecular graphs and measuring the degree of similarity between different graphs.
In this work, we address these challenges by extending our recently developed AD approach based on one-class support vector machines (SVMs)~\citep{Schweidtmann2021_AD} to GNNs.
A one-class SVM is a ML model that can be used to identify outliers by classifying whether an input is similar or dissimilar to the training data.
We train one-class SVMs on the molecular fingerprint of the GNN (cf. Figure~\ref{fig:GNNstructure}) to determine the GNN's AD.
We then restrict our molecular design loop to molecules which are accepted by the SVM (cf. Equation~\eqref{eq:RONOS_constrOpt}) which formally means $\mli{AD}(\cdot) = \mli{SVM}_{\text{AD}}(\mathbf{h}_{\text{FP,train}})~\geq~0$ where $\mathbf{h}_{\text{FP,train}}$ is the molecular fingerprint computed by the GNN and $\mli{SVM}_{\text{AD}}$ denotes the trained SVM.
The underlying idea for the AD is that the GNN computes similar molecular fingerprints whenever two molecules are structurally similar.
Since our prediction model is an ensemble of multiple GNNs, we train one SVM for each GNN model in the ensemble and apply a majority vote.
That is, each SVM $j$ evaluates $\mli{SVM}_{\text{AD},j}(\mathbf{h}_{\text{FP}})$ and returns $1$ if the molecule lies within the AD or $-1$ if not.
Subsequently, we sum up the votes to decide if the prediction of the GNN ensemble (EL) for a new molecules is classified as reliable, i.e., $\mli{SVM}_{\text{AD-EL}}(\mathbf{h}_{\text{FP}})~=~\sum_{j}~\mli{SVM}_{\text{AD},j}(\mathbf{h}_{\text{FP}})~\overset{!}{>}~0$.
Note that further details on the AD are described in the ESI.

\subsection{Implementation and hyperparameters}\label{subsec:ImplementationHyperparameter}
We implement our graph-ML CAMD framework in Python with the cheminformatic package RDKit~\citep{rdkit} and the ML frameworks pytorch and tensorflow, accounting for the different implementations of the generators, and provide our code open-source, see~\citep{Graph_ML_Fuel_Design_GIT}.
Moreover, we follow the implementation of the MHG-VAE by Kajino~\citep{Kajino2019} and use Luigi~\citep{Luigi2012} to automate computational experiments.
For the three generators, JT-VAE~\citep{Jin2018}, MHG-VAE~\citep{Kajino2019}, and MolGAN~\citep{DeCao2018}, we use the original implementations and hyperparameters as provided in the respective study and code repository and only extend the code to work in our framework.
We train the molecule generation models on all HCO-molecules in the QM9 data set~\citep{Ruddigkeit2012, Ramakrishnan2014}, i.e., all molecules within QM9 that contain exclusively hydrogen, carbon, and oxygen atoms.
QM9 contains approximately 50,000 HCO-molecules from various molecular classes.
We use the original implementation and model parameters of our GNN~\citep{Schweidtmann2020_GNNs} which is based on pytorch-geometric~\citep{Fey2019_PyGeo}.
The SVMs for the AD are implemented with scikit-learn~\citep{scikit-learn} building on our AD study~\citep{Schweidtmann2021_AD}.
For BO, we use GPyOpt~\citep{gpyopt2016}.
Note that we did not attempt deterministic global optimization of the acquisition function within the BO, e.g., by using our tool MeLOn~\citep{schweidtmann2019_GOANN, Schweidtmann2021_GOGP}, due to the high dimensionality (cf. Subsection ``Optimization'') and associated high computational cost. 
Thus, we use the local optimization algorithm L-BFGS~\citep{Liu.1989} implemented in GPyOpt~\citep{gpyopt2016}.
As GA, we use the python package geneticalgorithm~\citep{GApip2020}.
For both BO and GA, we apply default settings.
We follow the study of MHG-VAE by Kajino~\citep{Kajino2019} and reduce the dimensionality of the latent space within the VAEs by means of PCA aiming for an explained variance ratio of 99.9\% (JT-VAE: from 56 to 41, MHG-VAE: from 72 to 38) before performing BO.
Further details on the hyperparameter choice can be found in the ESI.
We run all computations on the HPC-cluster (CLAIX-2018) of RWTH Aachen University using one Supermicro 1029GQ-TVRT-01 node of an Intel Platinum 8160 core with 192 GB RAM, of which we used at most 8 GB, plus one NVIDIA Volta V100-SXM2 16 GB GPU.
For reproducibility, we fixed random seeds for training the models and running the design loop that we provide with our code.

\section{Results and discussion}\label{sec:Results}
We first present the computational results of our graph-based CAMD of high-octane fuels and then provide a discussion of the top candidates to demonstrate both strengths and potential weaknesses of the fully data-driven design approach.

\subsection{CAMD results}\label{subsec:CAMDresults}

We test all combinations of the three generator models (JT-VAE, MHG-VAE, and MolGAN) and the two optimization approaches (BO and GA) as well as two different stopping criteria (SC), i.e., a limit on the number of candidate molecules generated (SC$_\text{\#molecs}$) and an upper limit on the wall-clock run time (SC$_\text{time}$).
For SC$_\text{\#molecs}$, we consider both the number of unique molecules (1,000) and the total number of molecules (2,000) generated, as the number of duplicates can otherwise cause an unlimited run time.
In the SC$_\text{\#molecs}$ setting, the design loop will typically run for 0.5 to 8 hours.
The run time limit in SC$_\text{time}$ is set to 12 hours to investigate the effects of keeping the design loop running for a longer time.
Furthermore, we distinguish between runs with and without the AD.
All design loop runs are run five times (initialized with different random seeds) and the results are aggregated.

\begin{figure*}
	\begin{subfigure}[c]{0.3\textwidth}
		\centering
		\includegraphics[width=\textwidth]{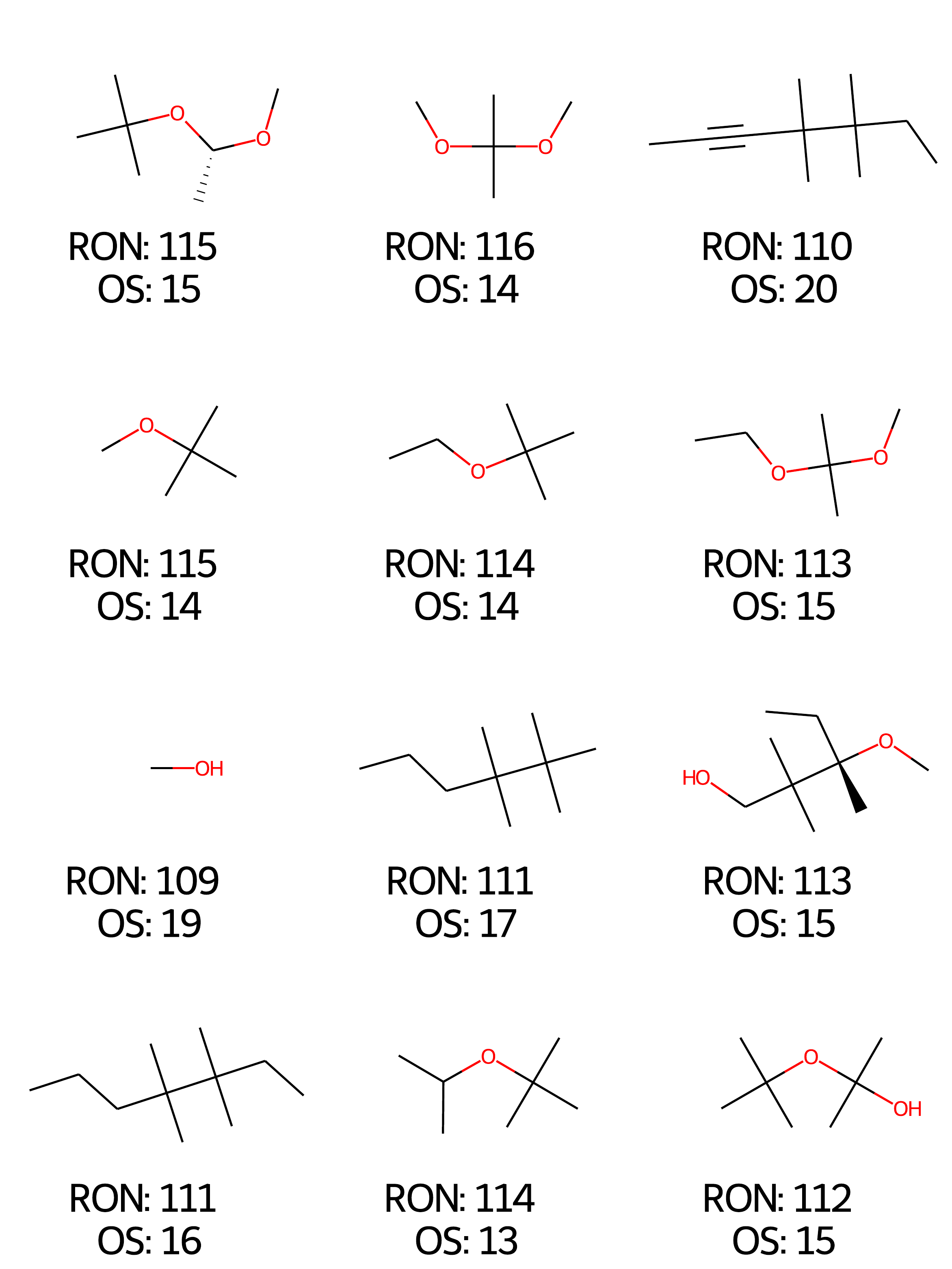}
		\subcaption{JT-VAE}
	\end{subfigure}
	\hfill
	\begin{subfigure}[c]{0.3\textwidth}
		\centering
		\includegraphics[width=\textwidth]{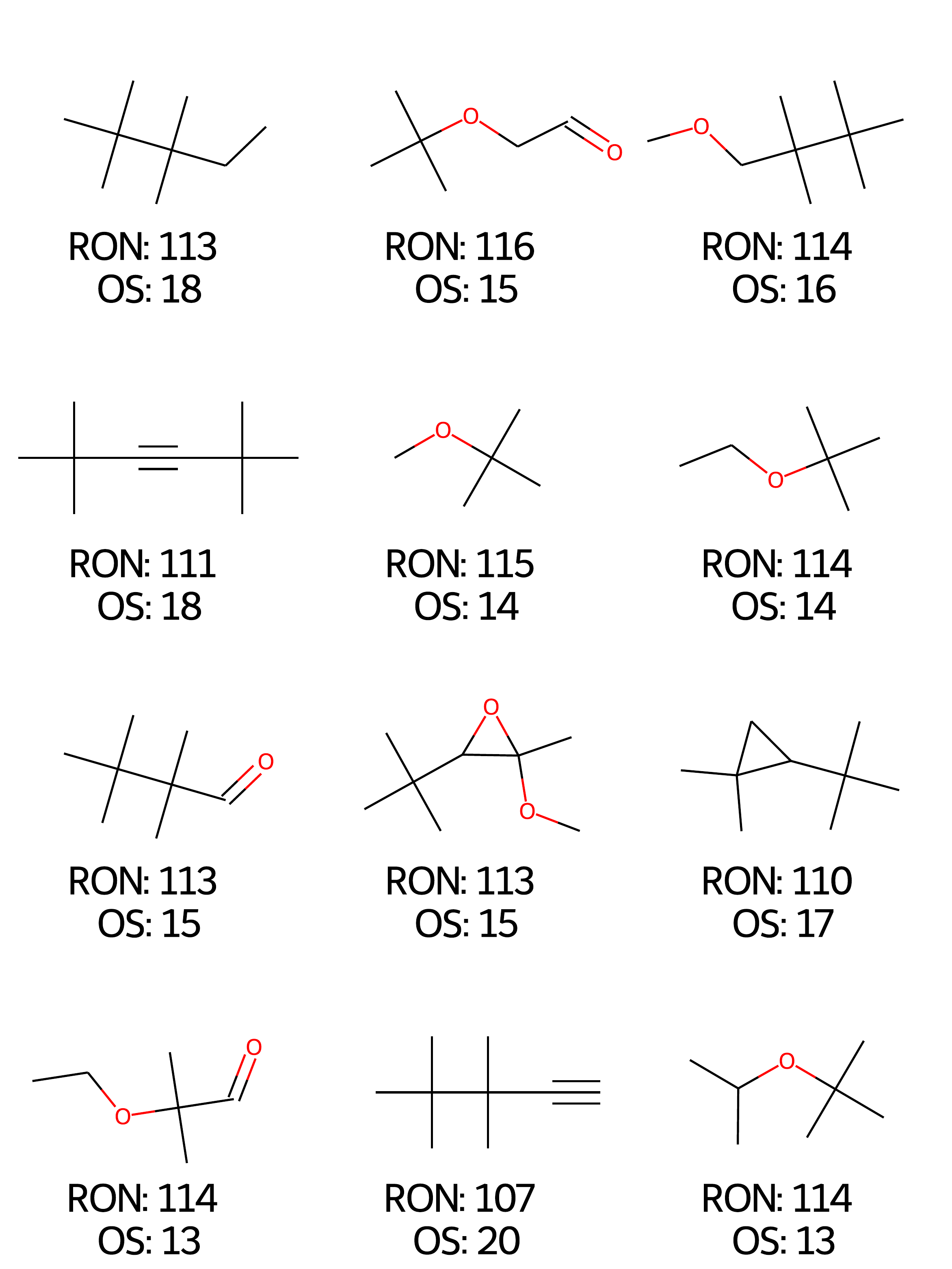}
		\subcaption{MHG-VAE}
	\end{subfigure}
	\hfill
	\begin{subfigure}[c]{0.3\textwidth}
		\centering
		\includegraphics[width=\textwidth]{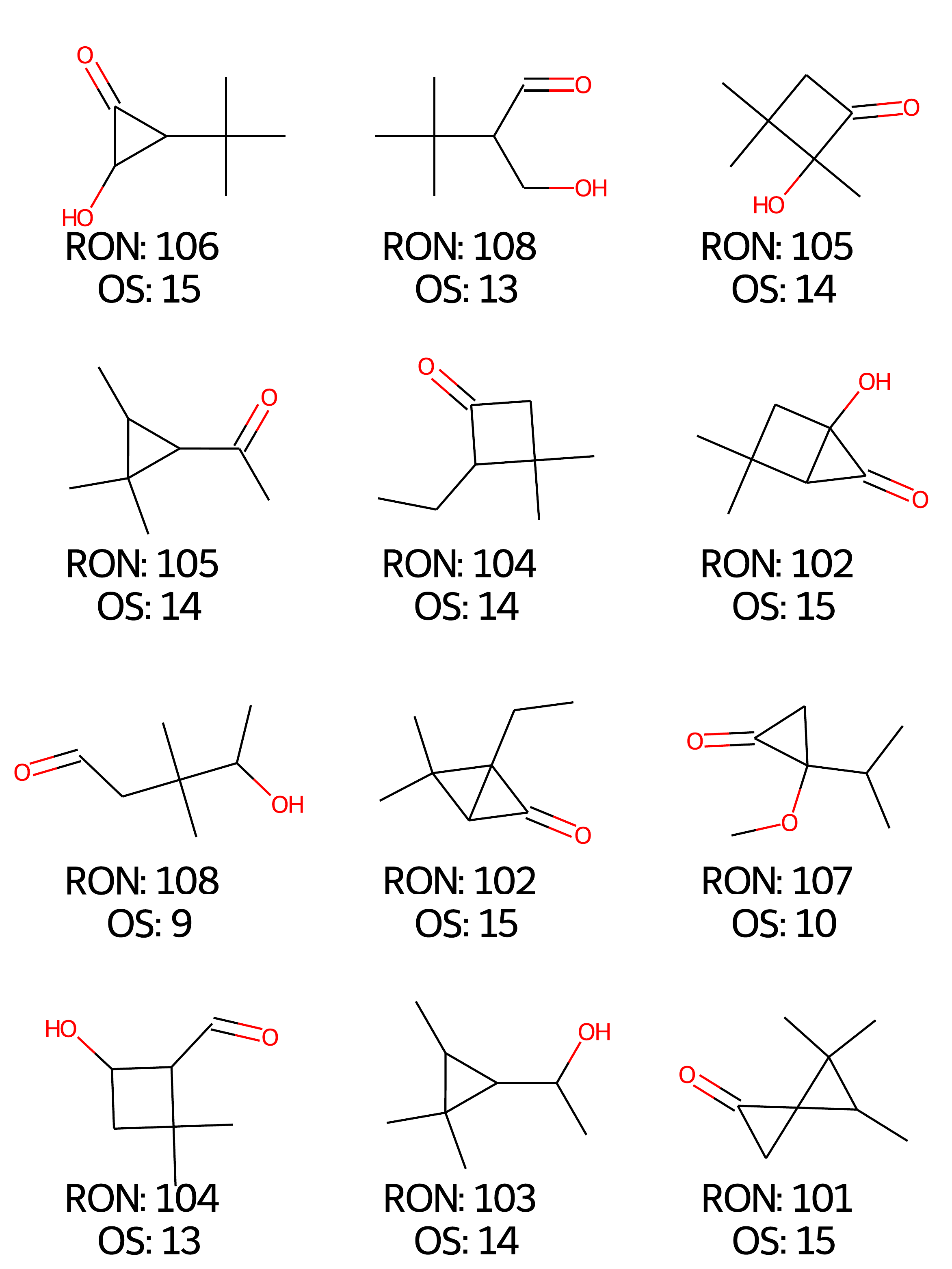}
		\subcaption{MolGAN}
	\end{subfigure}
	\caption{Top 12 candidates identified by the three different generator models with stopping criterion SC$_\text{\#molecs}$ (max. 1000 unique molecules or max. 2000 total molecules) and applicability domain. RON and OS values are predicted by the graph neural network~\citep{Schweidtmann2020_GNNs}.}
	\label{fig:TopMolsGenerators}
\end{figure*}

The top 12 molecules identified with SC$_\text{\#molecs}$ and active AD for the respective generators are shown in Figure~\ref{fig:TopMolsGenerators} together with the predicted RON and OS values.
The results demonstrate that the generators successfully propose molecules with high predicted RON and OS. 
Moreover, the top molecules are from a variety of different molecular classes, e.g., ethers, alcohols, and ketones, some of which are known to contain promising SI engine fuel candidates~\citep{Dahmen.2016}.
The majority of molecules has at least one oxygen atom.
Almost all top molecules generated by MolGAN include a cyclic structure, often associated with a cyclopropane feature, which we attribute to the high RON and OS for components with a cyclopropane substructure in the training set of the GNN model~\citep{Schweidtmann2020_GNNs_ESI}. 
Most top molecules generated by the two VAE models include strongly branched non-cyclic components, often in combination with one or two oxygen atoms, which are also known for high RON and OS values.
Both VAE models generate the popular octane enhancers MTBE and ETBE, and some related small, branched ether structures.
The JT-VAE also identifies ethanol, the prototype biofuel for SI engines.

\begin{table*}[t]
	\centering
	\caption{Results of optimization over 5 runs each. A molecule is considered promising if both RON\,$>$\,110 and OS\,$>$\,10. Runs with applicability domain are indicated by +AD.}
	\label{tab:ResultTable_RONOS}%
	\resizebox{\linewidth}{!}{%
		\begin{tabular}{cl|cccc|cccc|cccc}
			\toprule
			\multicolumn{2}{c}{\multirow{2}[3]{*}{\shortstack[c]{predicted RON$+$OS}}} & \multicolumn{4}{c}{JT-VAE} & \multicolumn{4}{c}{MHG-VAE} & \multicolumn{4}{c}{MolGAN} \\
			\cmidrule(lr){3-6}\cmidrule(lr){7-10}\cmidrule(lr){11-14}
			\multicolumn{2}{c|}{}   & BO & BO+AD & GA & GA+AD & BO & BO+AD & GA & GA+AD & BO & BO+AD & GA & GA+AD \\
			\midrule
			\multicolumn{1}{c}{\multirow{4}[2]{*}{\untereinander{SC$_\text{\#molecs}$}{\untereinander{(1000 unique molecules,}{\untereinander{2000 total)}{\#runs: 5}}}}} 
			& max               & 205 & 130 & 129 & 130 & 138 & 129 & 136 & 131 & 121 & 121 & 121 & 121 \\
			& mean top 20       & 181 & 125 & 125 & 126 & 131 & 125 & 132 & 128 & 110 & 111 & 116 & 116 \\
			& \# unique mol.    & 2390 & 1347 & 3472 & 3712 & 4671 & 4308 & 4683 & 4427 & 21 & 21 & 46 & 46 \\
			& \# promising mol. & 117 & 10 & 15 & 19 & 45 & 9 & 52 & 30 & 0 & 0 & 0 & 0 \\
			\midrule
			\multicolumn{1}{c}{\multirow{4}[2]{*}{\untereinander{SC$_\text{time}$}{\untereinander{(12h run time)}{\#runs: 5}}}} 
			& max               & 205 & 130 & 187 & 131 & 138 & 129 & 145 & 131 & 121 & 121 & 121 & 121 \\
			& mean top 20       & 183 & 126 & 180 & 130 & 133 & 126 & 140 & 129 & 111 & 112 & 118 & 118 \\
			& \# unique mol.    & 2996 & 1935 & 109830 & 80818 & 6710 & 7081 & 55255 & 46989 & 22 & 23 & 193 & 172 \\
			& \# promising mol. & 140 & 12 & 2096 & 376 & 104 & 15 & 678 & 142 & 0 & 0 & 0 & 0 \\
			\bottomrule
		\end{tabular}%
	}
\end{table*}

\begin{figure}[htb]
	\begin{subfigure}[c]{0.47\textwidth}
		\centering
		\includegraphics[width=\textwidth]{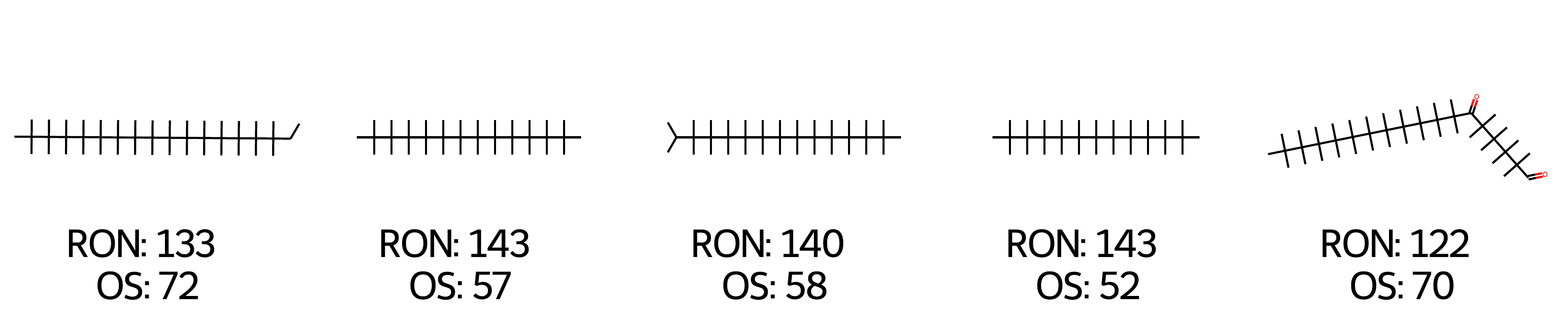}
		\subcaption{JT-VAE}
	\end{subfigure}
	\hfill
	\begin{subfigure}[c]{0.47\textwidth}
		\centering
		\includegraphics[width=\textwidth]{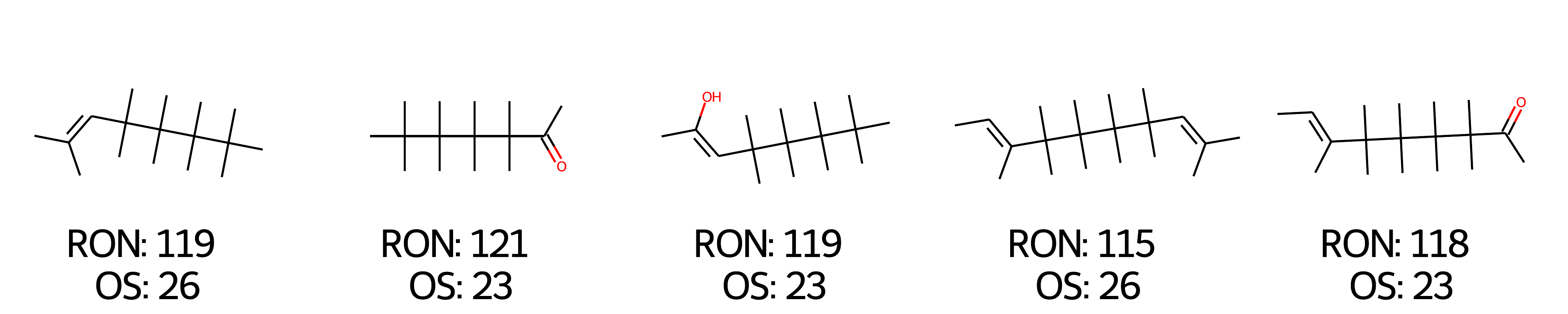}
		\subcaption{MHG-VAE}
	\end{subfigure}
	\caption{Top 5 candidates identified by the two VAE generator models with stopping criterion SC$_\text{time}$ (12 run time) and without applicability domain. All RON and OS values are GNN predictions~\citep{Schweidtmann2020_GNNs}.}
	\label{fig:MolsWithoutAD}
\end{figure}

Table~\ref{tab:ResultTable_RONOS} shows the statistics of all the runs with and without the AD, whereby each entry corresponds to the aggregated results over five runs.
Both the maximum and the mean predicted \ronos{} are typically lower if the AD is used.
In most cases, also the total number of molecules generated is lower if the AD is considered.
The observation that the AD often reduces the exploration performance is expected and in fact intended as the AD prohibits the generators from exploring structures that are far from the training data by strongly extrapolating the GNN model.
We want to emphasize that we find the generators to mainly produce chemically valid molecules.
Otherwise, e.g., MolGAN sometimes generates disconnected substructures, the generated molecule is dropped so that effectively no chemically invalid structures are provided to the GNN and AD.
Note that generated molecules, which are considered highly dissimilar to the training molecules by the AD, can still be chemically valid.
We show examples of such chemically valid molecules well outside the GNN's AD in Figure~\ref{fig:MolsWithoutAD}, where the top candidates identified by the two VAEs with SC$_\text{time}$ are depicted; we refer to the ESI for further examples.

When visually inspecting the top molecules from the design runs without AD, we find that the obtained molecules are typically huge, strongly branched hydrocarbons, e.g., with up to almost 50 carbon atoms.
As such compounds are presumably solid at room temperature, they are not suitable as fuels.
To avoid solid molecules, a constraint on the melting point could be included in the design loop.
However, the melting point can only serve as a rough proxy for the suitability of a compound as an octane booster, since miscibility and volatility also depend on the composition of the base fuel and the blending ratio~\citep{McCormick.2017, Konig.2020}.
Some of the proposed large molecules might be soluble in a fuel blend, which could be evaluated in further investigations of mixture properties, but is beyond the scope of this work.
Furthermore, the RON and OS predictions for the molecules identified with the JT-VAE without AD (cf. Figure~\ref{fig:MolsWithoutAD}a) are visibly higher than the maximum RON (of 120 for 1,3,5-trimethylbenzene~\citep{Schweidtmann2020_GNNs, Derfer.1958}) and the maximum OS (of 36 for 1,4-cyclohexadiene~\citep{Schweidtmann2020_GNNs, Derfer.1958}) of the data used to train the GNN prediction model, indicating strong extrapolation.
In the following, we therefore present and discuss only those results that have been obtained with the AD.

We observe that the VAE generators predict molecules with a maximum \ronos{} of about 130 while MolGAN achieves a maximum of only 121 (cf. Table~\ref{tab:ResultTable_RONOS}).
The maximum \ronos{} values of slightly above 130 for the two VAE models are in good agreement with known high-octane fuels such as MTBE with its experimentally validated \ronos{} of 135.
The encouraging performance of both VAE generators thus shows the general feasibility of our graph-ML CAMD framework utilizing the SVM-based AD.

To further compare the different generator and optimization combinations, we analyze the number of distinct molecules generated as well as the number of molecules with promising ignition properties, i.e., the molecules with both a predicted RON\,$>$\,110 and a predicted OS\,$>$\,10.
Both VAEs find a large number of distinct molecules irrespective of the employed stopping criteria (cf. Table~\ref{tab:ResultTable_RONOS}).
Specifically for SC$_\text{\#molecs}$, both VAEs generate more than 3,500 unique molecules out of 5,000 maximally possible unique molecules (1,000 unique molecules each over 5 runs).
This means that not only do the VAEs find a large number of distinct molecules in each run, but the identified molecules also vary greatly between different runs, thus leading to an overall small number of duplicates.
In contrast, MolGAN mainly generates duplicates of which none are considered promising (cf. Table~\ref{tab:ResultTable_RONOS}).
Comparing the results for SC$_\text{\#molecs}$ and SC$_\text{time}$ (cf. Table~\ref{tab:ResultTable_RONOS}), it can be seen that the VAE-GA combinations significantly increase the number of both explored and promising candidates with longer run time.
Apparently, this observation does not extend to BO, with one possible explanation being that BO becomes inherently slower as more data points are added to the surrogate model, thereby reducing the number of predictions per time, whereas the corresponding rate remains unchanged in the GA (cf. Subsection ``Optimization'').

\begin{figure*}
	\centering
	\captionsetup[subfigure]{justification=centering}
	\begin{subfigure}[c]{0.5\textwidth}
		\centering
		\includegraphics[width=\textwidth]{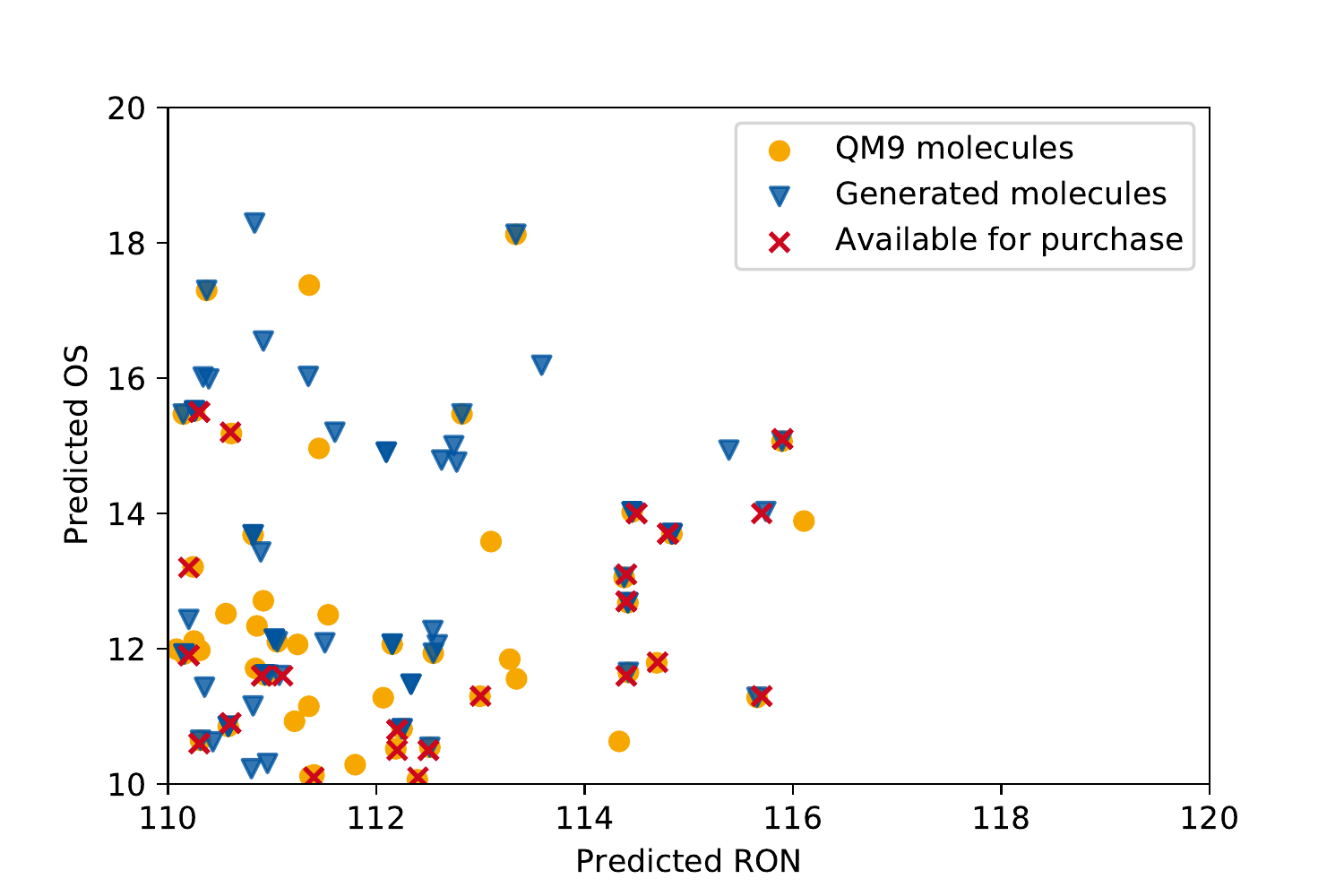}
		\subcaption{SC$_\text{\#molecs}$ (1000 unique molecules, \\ 2000 total), all generators, with AD}
	\end{subfigure}%
	\begin{subfigure}[c]{0.5\textwidth}
		\centering
		\includegraphics[width=\textwidth]{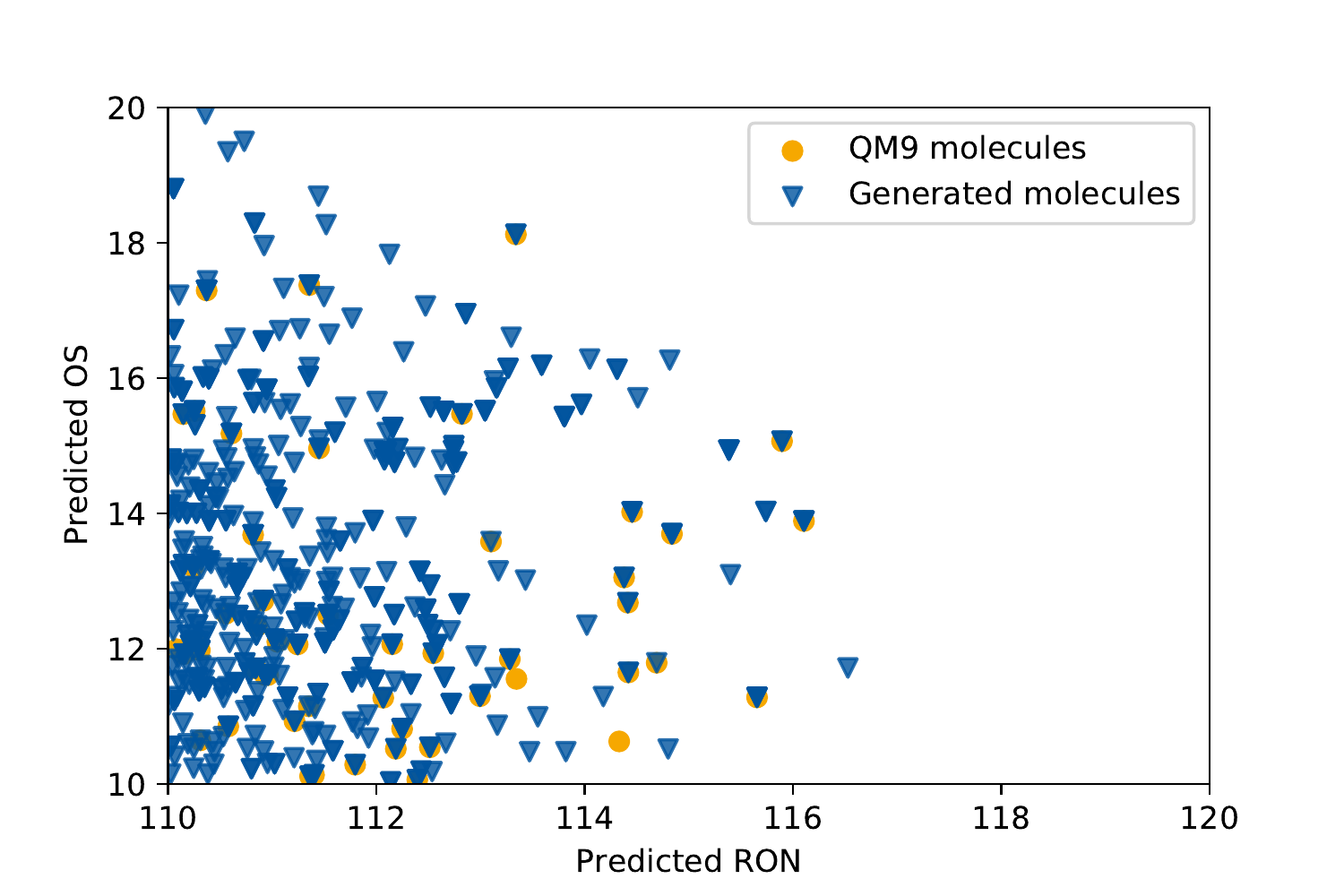}
		\subcaption{SC$_\text{time}$ (12h run time),\\all generators, with AD}
	\end{subfigure}
	\caption{Promising candidates (predicted RON\,$>$\,110 and OS\,$>$\,10). Commercially availability (red crosses) determined by manual search on Sigma-Aldrich and Chemspider websites~\citep{SigmaAldrich, Chemspider}.}
	\label{fig:RONvsOS}
\end{figure*}

The predicted RON and OS values of all promising molecules obtained with the two stopping criteria are shown in Figure~\ref{fig:RONvsOS}.
We also highlight those molecules identified in the SC$_\text{\#molecs}$ setting that are commercially available at chemical suppliers.
Commercial availability was assessed by a manual search on Sigma-Aldrich~\citep{SigmaAldrich} and Chemspider~\citep{Chemspider} websites without imposing a price limit but only including those molecules with an explicitly stated price; we did not search for the lowest price on different websites.
For SC$_\text{time}$, Figure~\ref{fig:RONvsOS}b, the effort for a manual search was considered disproportional due to the high number of promising candidates.
We further indicate molecules with high predicted \ronos{} in the QM9 database~\citep{Ruddigkeit2012, Ramakrishnan2014} that is used for training the generative models; additional QM9 statistics are provided in the ESI.
Figure~\ref{fig:RONvsOS} demonstrates that the graph-ML CAMD framework is able to generate molecules with high predicted RON and high predicted OS that are not in the QM9 database.
This observation is emphasized in case of SC$_\text{time}$ (cf. Figure~\ref{fig:RONvsOS}b). 
The capabilities of the generator models to generalize therefore allow to explore novel molecules for further investigation.

\subsection{Discussion of top candidates}\label{subsec:DiscTopCand}
In the discussion of the top molecules, we restrict our analysis to the promising molecules (RON\,$>$\,110 and OS\,$>$\,10 ) generated using SC$_\text{\#molecs}$, as the number of molecules generated with SC$_\text{time}$ is very large; we refer to the ESI for a detailed list of all generated promising molecules.
The top molecules that are also commercially available are illustrated in Table~\ref{tab:SCi_PromisingMolsToBuy}, including RON and OS predictions, literature values for RON and OS (where available), price category, and the respective combinations of generator and optimizer that identified the molecule.

\begin{table*}[htbp]
	\renewcommand{\arraystretch}{2}
	\centering
	\caption{All 16 commercially available molecules with predicted RON\,$>$\,110 and OS\,$>$\,10 (identified in SC$_\text{\#molecs}$ setting and active applicability domain). 
		RON and OS data available in the literature are stated in parentheses.
		Prices are categorized based on data from different chemical suppliers~\citep{SigmaAldrich, Chemspace, Enamine, Synquestlabs}: $\leq 1000$\$/l (low), $> 1000$ \$/l and $\leq 10\,000$ \$/l (medium), $> 10\,000$ \$/l (high).}
	\label{tab:SCi_PromisingMolsToBuy}%
	\resizebox{\linewidth}{!}{%
		\begin{tabular}{c|cc|cc|c|c}
			\toprule
			\textbf{Class} & \textbf{Structure} & \textbf{SMILES} & \textbf{RON} & \textbf{OS} & \textbf{Price category} & \textbf{Generator (Optimizer)} \\
			\midrule 
			%
			\multirow{2}{*}[-10pt]{alkanes} 
			& \includegraphics[valign=c,height=0.5cm]{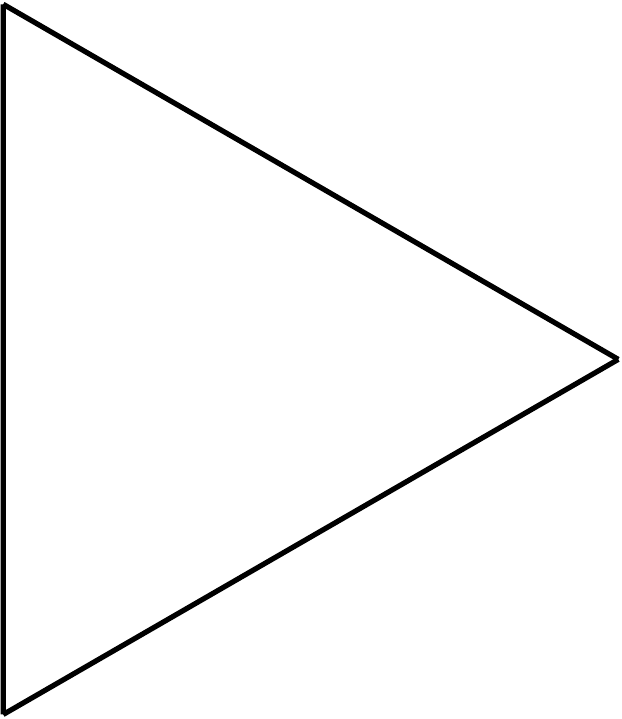} & \untereinanderstretch{C1CC1}{cyclopropane}                 & 110 & 16 & medium & \untereinanderstretch{JT (BO, GA),}{MHG (BO, GA)} \\
			& \includegraphics[valign=c,valign=c,height=1cm]{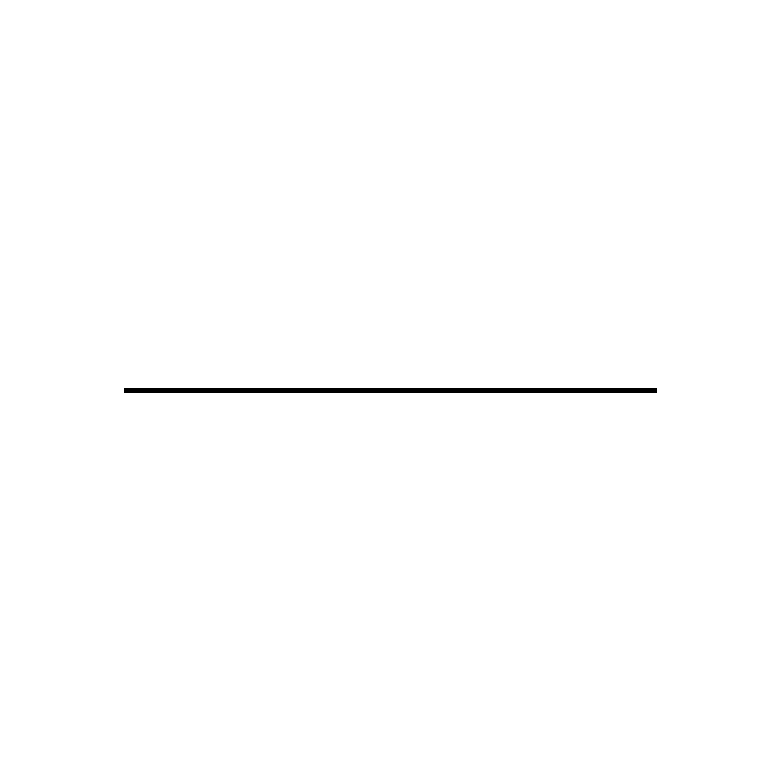} & \untereinanderstretch{CC}{ethane}                    & \untereinanderstretch{110}{(111~\citep{Derfer.1958})} & \untereinanderstretch{12}{(11~\citep{Derfer.1958})} & low & JT (BO, GA)  \\
			\midrule
			\multirow{1}[2]{*}{aromatics}
			& \includegraphics[valign=c,height=0.5cm]{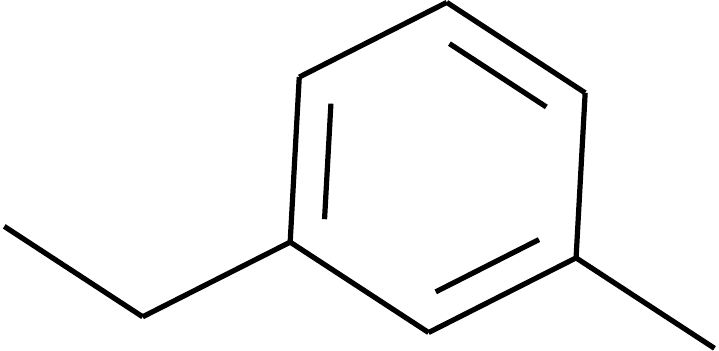} & \untereinanderstretch{CCc1cccc(C)c1}{3-ethyltoluene}           & \untereinanderstretch{110}{(112~\citep{Derfer.1958})} & \untereinanderstretch{11}{(12~\citep{Derfer.1958})} & high & JT (GA) \\
			\midrule
			%
			%
			\multirow{3}{*}[-20pt]{ethers} 
			& \includegraphics[valign=c,height=0.5cm]{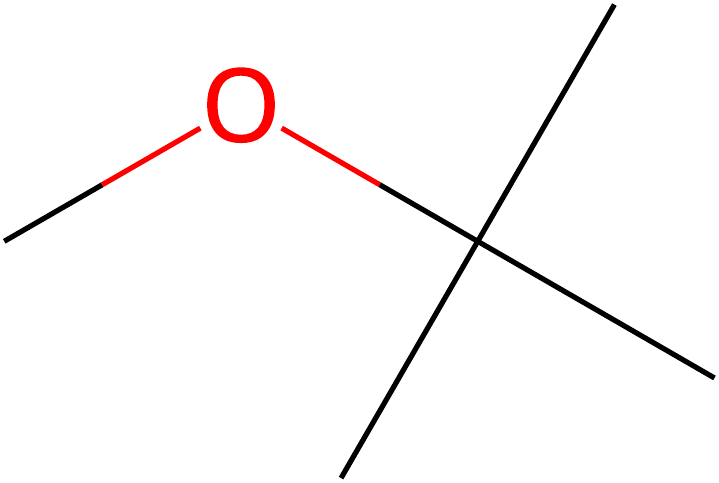} & \untereinanderstretch{COC(C)(C)C}{MTBE}            & \untereinanderstretch{115}{(118~\citep{leppard1991autoignition})} & \untereinanderstretch{14}{(17~\citep{leppard1991autoignition})} & low & \untereinanderstretch{JT (BO, GA),}{MHG (BO, GA)} \\
			& \includegraphics[valign=c,height=0.5cm]{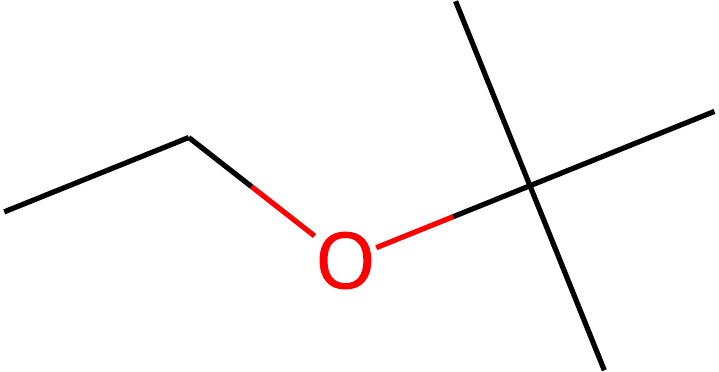} & \untereinanderstretch{CCOC(C)(C)C}{ETBE}            & \untereinanderstretch{114}{(118~\citep{Kubic.2017})} & \untereinanderstretch{14}{(16~\citep{Kubic.2017})} & medium & \untereinanderstretch{JT (BO, GA),}{MHG (GA)} \\
			& \includegraphics[valign=c,height=0.5cm]{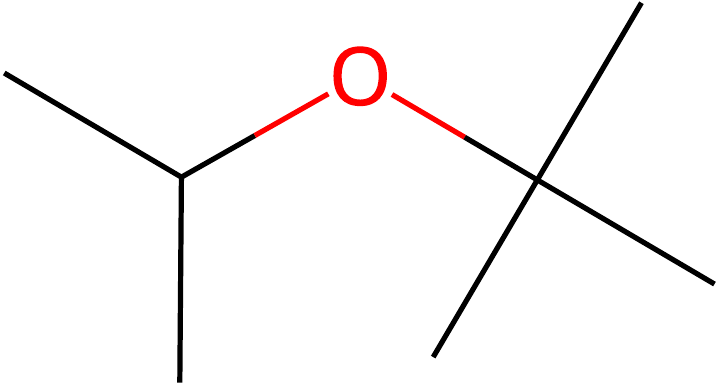} & \untereinanderstretch{CC(C)OC(C)(C)C}{tert-butyl isopropyl ether}       & 114 & 13 & high & \untereinanderstretch{JT (GA),}{MHG (GA)} \\
			\midrule
			\multirow{2}{*}[-10pt]{aldehydes} 
			& \includegraphics[valign=c,height=0.5cm]{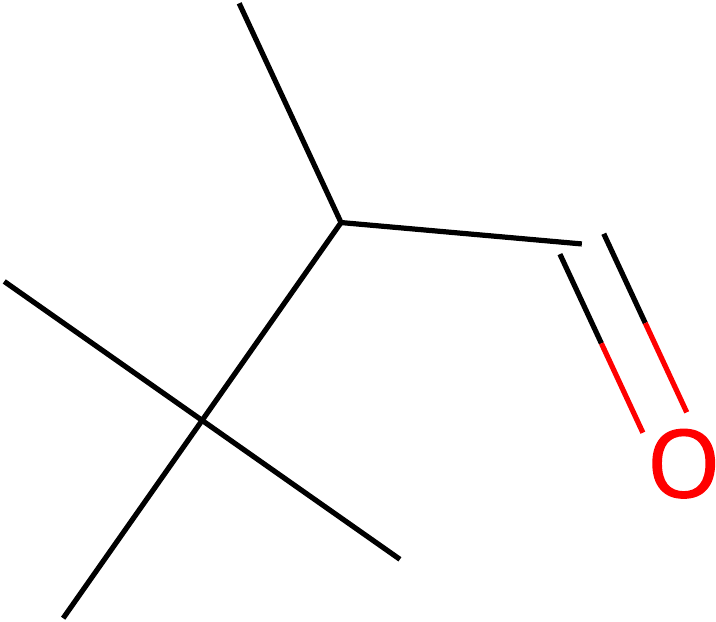} & \untereinanderstretch{CC(C=O)C(C)(C)C}{2,3,3-trimethylbutanal}   & 111 & 12 & high & MHG (GA) \\
			& \includegraphics[valign=c,height=0.5cm]{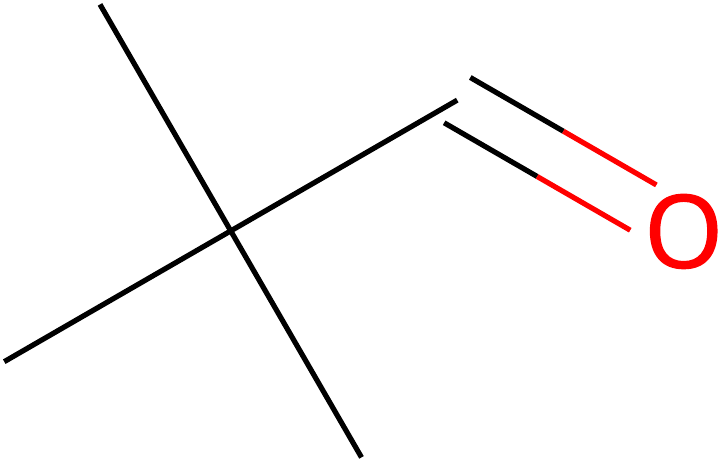} & \untereinanderstretch{CC(C)(C)C=O}{trimethylacetaldehyde}           & 111 & 11 & medium & \untereinanderstretch{JT (GA),}{MHG (BO)} \\
			\midrule
			\multirow{5}{*}[-40pt]{\untereinanderstretch{polyfunctional}{(aldehyde + ether)}} 
			& \includegraphics[valign=c,height=0.5cm]{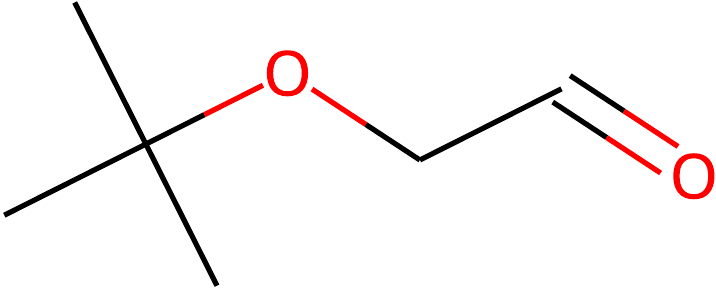} & \untereinanderstretch{CC(C)(C)OCC=O}{tert-butoxyacetaldehyde}      & 116 & 15 & high & MHG (GA) \\
			& \includegraphics[valign=c,height=0.5cm]{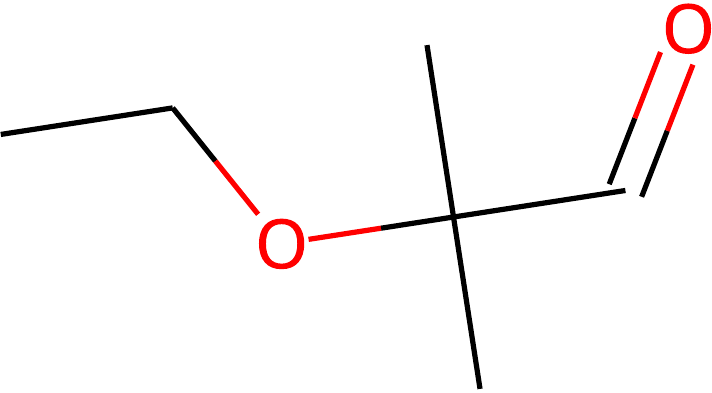} & \untereinanderstretch{CCOC(C)(C)C=O}{2-ethoxy-2-methylpropanal}          & 114 & 13 & high & MHG (GA) \\
			& \includegraphics[valign=c,height=0.5cm]{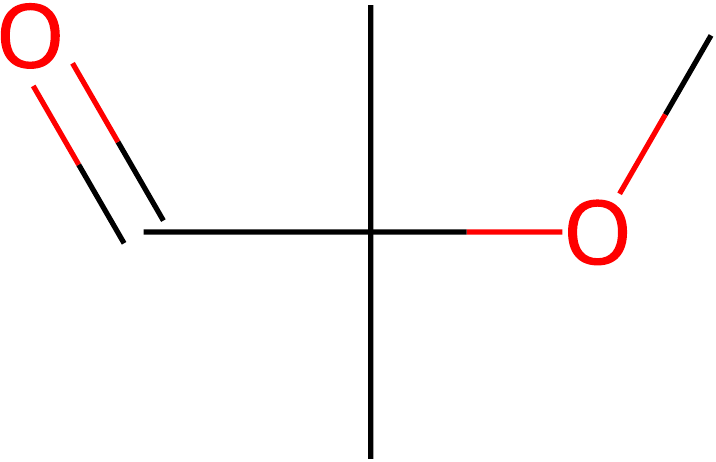} & \untereinanderstretch{COC(C)(C)C=O}{2-methoxy-2-methylpropanal}         & 116 & 11 & high & MHG (GA) \\
			& \includegraphics[valign=c,height=0.5cm]{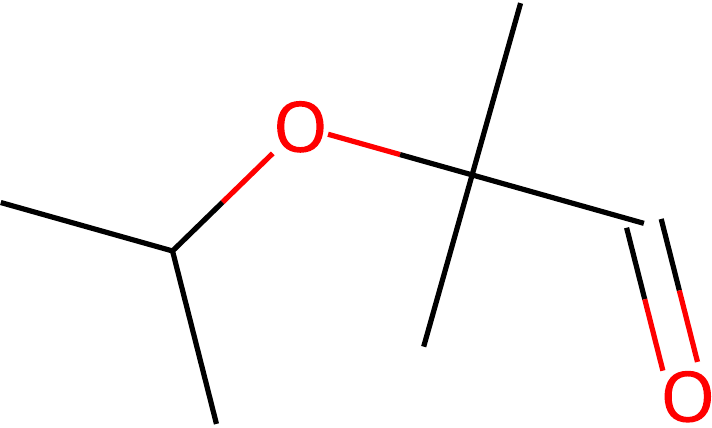} & \untereinanderstretch{CC(C)OC(C)(C)C=O}{2-methyl-2-propan-2-yloxypropanal}     & 114 & 12 & high & MHG (GA) \\
			& \includegraphics[valign=c,height=0.5cm]{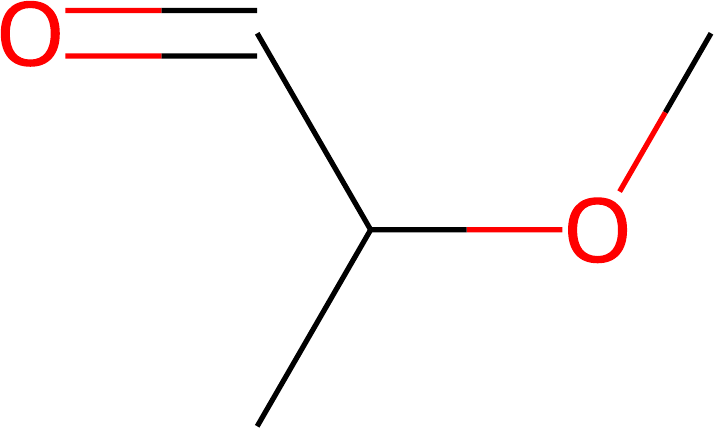} & \untereinanderstretch{COC(C)C=O}{2-methoxypropanal}              & 112 & 11 & high & MHG (BO, GA) \\
			\midrule
			\multirow{2}{*}[-10pt]{\untereinanderstretch{polyfunctional}{(ketone + ether)}} 
			& \includegraphics[valign=c,height=0.5cm]{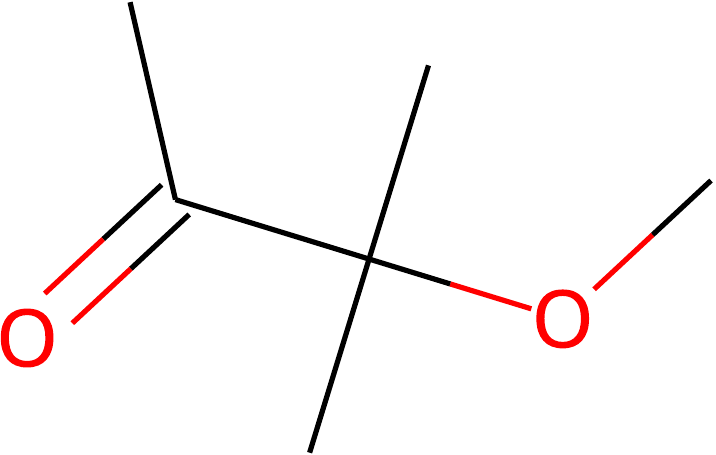} & \untereinanderstretch{COC(C)(C)C(C)=O}{3-methoxy-3-methyl-2-butanone}       & 113 & 11 & high & MHG (GA) \\
			& \includegraphics[valign=c,height=0.5cm]{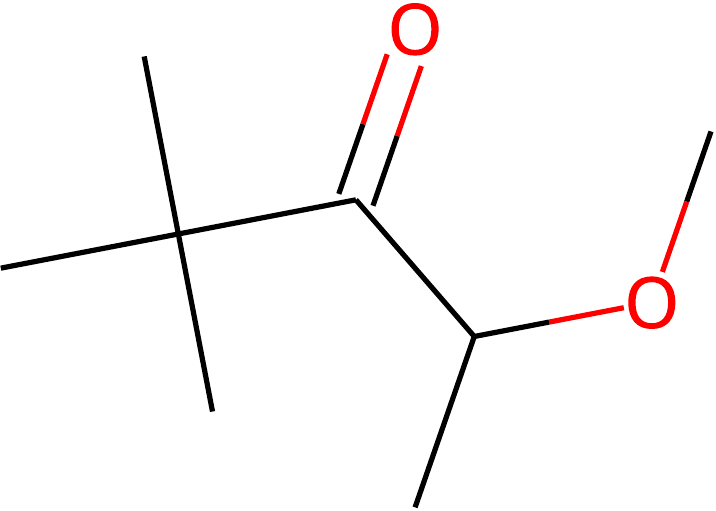} & \untereinanderstretch{COC(C)C(=O)C(C)(C)C}{4-methoxy-2,2-dimethylpentan-3-one}   & 111 & 12 & high & MHG (GA) \\
			\midrule
			\multirow{1}{*}[-5pt]{acetals} 
			& \includegraphics[valign=c,height=0.5cm]{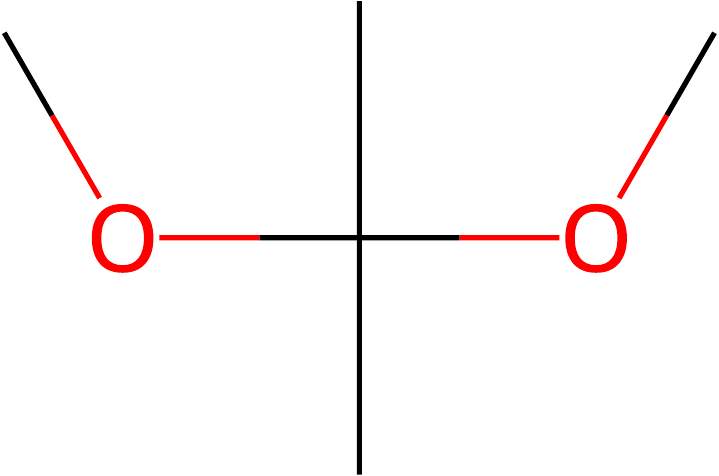} & \untereinanderstretch{COC(C)(C)OC}{2,2-dimethoxypropane}           & 116 & 14 & low & JT (GA) \\
			\bottomrule
		\end{tabular}%
	}
\end{table*}%

\subsubsection{Promising classes of molecules}\label{subsubsec:PromMolClass}
We find both pure hydrocarbons and oxygenated hydrocarbons (cf. Table~\ref{tab:SCi_PromisingMolsToBuy}), molecules already in use as octane boosters and molecules that constitute interesting candidates for further experimental investigation.
The two identified alkanes, ethane and cyclopropane, are gaseous under ambient conditions, whereas the one aromatic hydrocarbon, 3-ethyltoluene, is liquid.
The known \ronos{} scores from literature for ethane and 3-ethyltoluene of 122 and 124, respectively, are in good agreement with the GNN predictions.
We want to emphasize that gaseous compounds, such as ethane and cyclopropane, are difficult to implement as octane boosters.
To prevent gases within the candidate list, one could include boiling point constraints in the design loop.
However, the normal boiling point is, similar to the melting point discussed at the beginning of this section, only a rough preselection criterion, since the miscibility and volatility of a potential octane booster in a fuel blend strongly depend on the overall blend composition.
Next to alkanes, three ethers are identified, including methyl tert-butyl ether (MTBE) and ethyl tert-butyl ether (ETBE) that are used as octane boosters in practical applications~\citep{Demirbas.2015, Badia.2021}.
Their experimentally \ronos{} scores of 135 and 134~\citep{leppard1991autoignition, Kubic.2017} are slightly higher than the predicted scores.
Furthermore, molecules from the class of aldehydes are identified.
It has been found, however, that the formation of aldehydes during the combustion process of high-octane, oxygenated hydrocarbons results in increased exhaust emissions~\citep{Magnusson.2011}, indicating a lower suitability of aldehydes as fuels.
Polyfunctional molecules with an aldehyde and an ether group are generated as well, which also entail the problem of aldehyde emissions.
Further polyfunctional molecules containing an ether group and a ketone group are generated, with ketones being prominent high-octane fuels~\citep{Hoppe.2016_2butanone, Hechinger.2014}.
Most of the molecules containing an ether, a ketone, and/or an aldehyde functionality have a compact, branched structure with similarities to MTBE and ETBE, making them interesting high-octane fuel candidates; however, they also have a high price, hindering experimental investigation.

The last top candidate in Table~\ref{tab:SCi_PromisingMolsToBuy}, namely 2,2-dimethoxypropane (2,2-DMP), belongs to the class of acetales. 
It is a compact structure similar to ETBE, with the difference being that one carbon atom is replaced by a second oxygen atom. 
2,2-DMP also has a low price, making it an attractive target for experimental investigation.
A DCN measurement of 31 is known from literature~\citep{yanowitz2017compendium} which, however, is not suggestive of a very high RON, as molecules with RON $> 110$ typically correspond to DCN values below 10, cf.~\citep{Dahmen.2016, Perez.2012}.
Our high \ronos{} prediction (cf. Table~\ref{tab:SCi_PromisingMolsToBuy}), however, is consistent with the \ronos{} value of 143 stated in a recent study by Li et al.~\citep{Li.2022} who used a ML-QSPR prediction model combining both ML and a group contribution approach.
Another ML-based QSPR model for RON and OS recently developed by vom Lehm et al.~\citep{VomLehn.2020} likewise predicts a high \ronos{} value of 156.

\subsubsection{Comparison to previous fuel design studies}\label{subsubsec:CompCAMD}
Our commercially available top candidates (cf. Table~\ref{tab:SCi_PromisingMolsToBuy}) generally match the molecular classes identified in previous fuel design/screening studies for SI engine fuels, e.g., in~\citep{Dahmen.2016, Li.2022, Hoppe.2016, VomLehn.2021}.
Specifically, prominent molecular classes from previous studies include the herein identified groups of ethers~\citep{Dahmen.2016, Li.2022, VomLehn.2021}, ketones~\citep{Dahmen.2016, Li.2022, Hoppe.2016, VomLehn.2021}, aromatics~\citep{VomLehn.2021}, aldehydes~\citep{Li.2022, Hoppe.2016}, alkanes~\citep{VomLehn.2021}, and acetals~\citep{Li.2022}.
Interestingly, our top candidates do not include any esters, alcohols, and furans that have often been identified in the literature~\citep{Dahmen.2016, Li.2022, VomLehn.2021}.
When inspecting all molecules generated in our design loop runs with SC$_\text{\#molecs}$ and with AD, we indeed find esters (e.g., methyl acetate), alcohols (e.g., ethanol and methanol), as well as furans (e.g., 2-methylfuran).
However, these are not considered top candidates as predicted OS is below 10 for most esters and predicted RON is slightly below 110 in case of furans and alcohols.
Such RON and OS predictions are generally in accordance with the literature values for representative molecules of these classes, cf.~\citep{McCormick.2017, Schweidtmann2020_GNNs, Derfer.1958, Naegeli.1989, yanowitz2011utilization}.

The polyfunctional molecules identified in our study are hardly discussed in the literature.
It should be noted that the availability of experimental RON and MON values for polyfunctional molecules is very limited, indicating a high uncertainty in the GNN predictions.

The generated top candidate of acetals, 2,2-DMP, has also been identified in the fuel screening by Li et al.~\citep{Li.2022} and will be investigated experimentally in the following.

\subsubsection{Experimental assessment of 2,2-DMP}
\begin{figure}[htbp]
	\centering
	\includegraphics[width=0.5\textwidth, trim={1cm 9cm 1cm 9cm},clip]{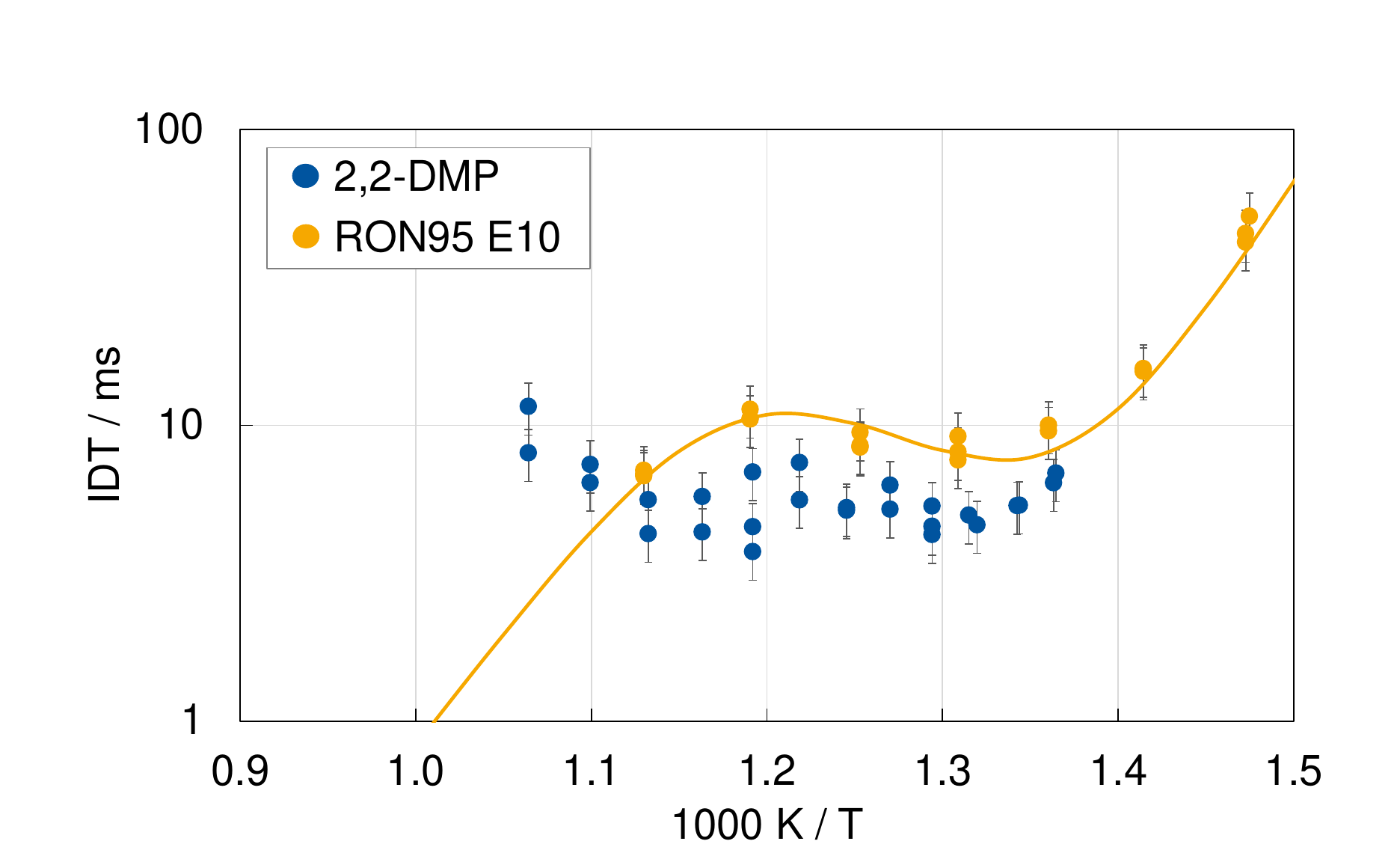}
	\caption{Measured ignition delay time in a rapid compression machine for 2,2-dimethoxypropane and a commercially available RON95$\,$E10 pump station fuel. The error bars indicate ±~20~\% scatter of the measured ignition delay time. The yellow line corresponds to a three-fold Arrhenius model fit to the RON95$\,$E10 ignition delay times~\citep{Weisser.2001}.}
	\label{fig:RCM_results}
\end{figure}

Experimental investigation of 2,2-DMP was conducted in dedicated test engines according to the DIN~EN~ISO~5164~\citep{DIN_EN_5164} and DIN~EN~ISO~5163 standards~\citep{DIN_EN_5163}, respectively, by an external company.
Measurement of RON and MON of pure 2,2-DMP, however, could not be performed.
Instead, blends of 2,2-DMP with 90~\%, 80~\%, and 60~\% (v/v) of gasoline were investigated. 
The extrapolation to pure component values yielded a RON of 91.75 (+/- 0.25) and a MON of 87.27 (+/- 0.3), hence a \ronos{} score of about 96, indicating a strong misprediction by our GNN model as well as the models by Li et al.~\citep{Li.2022} and by vom Lehn et al.~\citep{VomLehn.2020}.
To further clarify the ignition properties of 2,2-DMP, we experimentally measured ignition delay times (IDT) in a rapid compression machine (RCM)~\citep{LEE2012, RAMALINGAM2017} and
compared the chemical reactivity of 2,2-DMP to that of a typical RON95$\,$E10 pump station fuel.
IDT measurements for 2,2-DMP were performed at an end-of-compression pressure of 20 \si{bar} for a stoichiometric mixture and with an argon-to-oxygen dilution ratio of 3.762 in the temperature range of 733 to 940 \si{K}. 
Details on the RCM measurements can be found in the ESI.
The ignition took place via a two-stage process in the investigated temperature regime indicating strong low-temperature chemistry, cf. Figure~\ref{fig:RCM_results}, not representative for a high-octane fuel. 
Compared to the RON95$\,$E10 fuel, 2,2-DMP shows a distinctively higher reactivity between 740 and 870 \si{K} pointing towards a lower knock resistance and thus RON value. 
The RCM results suggest a slightly worse knock resistance of 2,2-DMP compared to RON95$\,$E10 pump station fuel, supporting the extrapolated RON and MON measurements.

The case of 2,2-DMP shows the potential weaknesses of a fully data-driven approach in a data-scarce environment.
We account the large model prediction error of our GNN as well as those of the models by Li et al.~\citep{Li.2022} and by vom Lehn et al.~\citep{VomLehn.2020} to the comparatively little training data available for RON and MON modeling.
Specifically, our RON and MON training database includes just five ethers, a single acetal (not 2,2-DMP), no aldehydes, eight ketones, and only two molecules with more than one type of oxygen functionality (cf.~\citep{Schweidtmann2020_GNNs}). 
Similar data limitations apply to the other RON and MON prediction models~\citep{Li.2022, VomLehn.2020}, explaining their similarly bad predictions in case of 2,2-DMP.
Furthermore, we want to stress the fact that no RON and MON values for aldehydes are included in the training data, so our GNN may not sufficiently distinguish between aldehydes and ketones.
The \ronos{} predictions of the identified molecules with an aldehyde group are therefore considered subject to large uncertainty.  
In the case of 2,2-DMP, a DCN data point was available and used in the training of our multi-task GNN for simultaneous RON, MON, and DCN prediction (cf. Subsection ``Property prediction'').
As expected, our AD approach based on majority voting (cf. Subsection ``Applicability domain'') considers 2,2-DMP within the region of reliable predictions as it was part of the training data.
Yet, only 31 out of 40 SVMs voted for 2,2-DMP.
Increasing the AD consensus level, e.g., 80\% instead of 50\%, may provide some protection against such strong mispredictions, at the cost of a smaller search space.
A systematic investigation of the relationship between the AD consensus level and the prediction accuracy for molecules proposed by the design loop, however, is beyond the scope of this work. 
The weak spots of prediction models for fuel ignition quality remain a huge challenge for model-based fuel design, even when utilizing state-of-the-art ML~\citep{Schweidtmann2020_GNNs, VomLehn.2020} and an applicability domain.
Therefore, acquiring more training data is absolutely crucial.

\section{Conclusion}\label{sec:Conclusion}

We propose a fully data-driven CAMD approach based on recent methods from graph-ML for the identification of molecules with desired ignition characteristics for modern SI engines.
Our graph-ML CAMD framework utilizes a representation of molecules as graphs and incorporates three modules for building a molecular design loop: (1) molecule generation from a continuous molecular space with generative graph-ML, (2) molecular property prediction through GNNs, and (3) optimization for strategic sampling from the continuous molecular space to find molecules with high predicted \ronos{}.
The modular structure enables the exploration of different ML models in combination with different optimization approaches.
We additionally present a novel approach to identify the applicability domain (AD) of GNN models for molecular property prediction.
By predicting promising high-octane fuel molecules in a fully data-driven fashion, our study exemplifies how recent developments in ML can be utilized for CAMD and its automation.

The top molecular candidates identified with our graph-ML CAMD framework are from well-known molecular classes for high-octane fuels, e.g., ethers and ketones, and include both well-established components like MTBE and ETBE as well as new promising candidates for further experimental investigation.
The comparison of different generative graph-ML models, namely JT-VAE~\citep{Jin2018}, MHG-VAE~\citep{Kajino2019}, and MolGAN~\citep{DeCao2018}, in combination with different optimization approaches, BO and GA, shows that the choice of the generative model and optimization strategy influences the number and type of identified candidate molecules.
Both VAEs provide a diverse continuous molecular space with a large number of potential molecules, while MolGAN generates a comparatively low number of candidates and yields lower target property values. 
We conclude that the GA is well suited for exploring large portions of the continuous molecular space of the generative models, especially when working with high dimensions where BO struggles but still finds some promising candidates.
Our AD approach additionally enables us to focus the exploration on candidates with presumably more accurate predictions.
The experimental investigation of one candidate within the AD, namely 2,2-dimethoxypropane, shows lower RON and OS values than predicted by our GNN model, demonstrating the limitations of CAMD in a comparatively data-scarce environment. 
We thereby highlight the importance of experimental validation to fuel design and the need for further RON and OS training data.
Furthermore, the correlation between the AD threshold, i.e., the consensus level, and the prediction accuracy for molecules proposed by the design loop should be investigated. 

Future work could include additional physical and chemical properties in the design, e.g., melting point, boiling point, vapor pressure, toxicity, or viscosity, similar to previous studies~\citep{Dahmen.2016, Li.2022, Hoppe.2016, VomLehn.2021}.
The framework, in principle, is not bound to fuel design as application but could also be applied to other CAMD applications such as drug discovery, design of catalysts, pesticides, etc.



\section*{Declaration of Competing Interest}
We have no conflict of competing interest.

\section*{Data Availability Statement}
The data that support the findings of this study are openly available in our GitLab repository ``Graph machine learning for design of high-octane fuels'' at \url{https://git.rwth-aachen.de/avt-svt/public/graph_ML_fuel_design}, reference number~\citep{Graph_ML_Fuel_Design_GIT}.

\section*{Acknowledgements}
This work was funded by the Deutsche Forschungsgemeinschaft (DFG, German Research Foundation) – 466417970 – within the Priority Programme ``SPP 2331: Machine Learning in Chemical Engineering''.
It was also funded by the Deutsche Forschungsgemeinschaft (DFG, German Research Foundation) under Germany’s Excellence Strategy - Cluster of Excellence 2186 ``The Fuel Science Center''.
This work was also performed as part of the Helmholtz School for Data Science in Life, Earth and Energy (HDS-LEE).
Further, this work was supported by the Deutsche Forschungsgemeinschaft (DFG, German Research Foundation) within the GRK 2236 UnRAVel. 
Simulations were performed with computing resources granted by RWTH Aachen University under project ``rwth0664''.
The authors thank Florian vom Lehn for providing RON and OS predictions for 2,2-dimethoxypropane by his model. 
MD received funding from the Helmholtz Association of German Research Centres.

\section*{Authors contributions}
\begin{itemize}[labelindent=0pt,labelwidth=0pt, labelsep*=0pt, leftmargin=!, style=standard]
	\item[] \textbf{Jan G. Rittig}: Conceptualization, Methodology, Software, Validation, Formal analysis, Investigation, Data curation, Writing - Original Draft, Writing - Review \& Editing, Visualization, Funding acquisition.	
	
	\item[] \textbf{Martin Ritzert}: Conceptualization, Methodology, Software, Validation, Formal analysis, Investigation, Data curation, Writing - Original Draft, Writing - Review \& Editing, Visualization.
	
	\item[] \textbf{Artur M. Schweidtmann}: Conceptualization, Methodology, Writing - Review \& Editing, Supervision, Funding acquisition. 
	
	\item[] \textbf{Stefanie Winkler}: Methodology, Validation, Software, Formal analysis, Data curation, Writing - Review \& Editing.		
	
	\item[] \textbf{Jana M. Weber}: Methodology, Writing - Review \& Editing.		
	
	\item[] \textbf{Philipp Morsch}: Investigation, Writing - Original Draft, Writing - Review \& Editing.	
	
	\item[] \textbf{K. Alexander Heufer}: Writing - Review \& Editing, Supervision, Funding acquisition.	
	
	\item[] \textbf{Martin Grohe}: Conceptualization, Writing - Review \& Editing, Supervision, Funding acquisition.
	
	\item[] \textbf{Alexander Mitsos}: Conceptualization, Writing - Review \& Editing, Supervision, Funding acquisition.
	
	\item[] \textbf{Manuel Dahmen}: Conceptualization, Formal analysis, Writing - Review \& Editing, Supervision.
\end{itemize}

  \clearpage
  \newpage

  \bibliographystyle{apalike}
  \renewcommand{\refname}{Bibliography}  
  \bibliography{literature.bib}

\end{document}


\twocolumn[
\begin{@twocolumnfalse}
	\thispagestyle{firststyle}
	
	\begin{center}
		\begin{large}
			\textbf{\mytitle}
		\end{large} \\
	\vspace{0.2cm}
		\myauthor
	\end{center}
	
	\vspace{-0.2cm}
	
	\begin{footnotesize}
		\affil
	\end{footnotesize}

	\vspace{0.9cm}
\end{@twocolumnfalse}
]

  \newpage

\section{Applicability domain for graph neural networks}\label{ESI:GNN_AD}
We apply our recently published applicability domain (AD) approach~\citep{Schweidtmann2021_AD} to graph neural networks (GNNs).
Specifically, we determine the AD of a GNN in the GNN's molecular fingerprint space by using a one-class support vector machine (SVM) to identify those molecular fingerprints that correspond to molecules for which our GNN tool presumably provides reliable predictions.
Using the molecular fingerprint vectors means that the SVMs are trained on fixed-size inputs, circumventing the need for handling the varying input size of molecular graphs.
Note that the dimension of the molecular fingerprint vector is a hyperparameter of the GNN and thus fixed before the training of the GNN.

For the training step of the AD, we extract the molecular fingerprints of the training molecules of the GNN and use those fingerprints as the training set for the one-class SVM.
We use an one-class SVM to determine the AD since we only have positive training samples, i.e., the fingerprints of the molecules we used for training the GNN.
After training, the SVM classifies predictions as non-reliable for every molecule that is not similar to the training data.
Technically, we select a linear classifier $\mli{SVM}_{\text{AD}}$ with $\mli{SVM}_{\text{AD}}(\mathbf{h}_{\text{FP,train}})~\geq~0$ where $\mathbf{h}_\text{FP,train}$ is a molecular fingerprint belonging to a molecule with its graph $G_\text{mol,train}$ seen during training, i.e., $\mathbf{h}_\text{FP,train} = g_\text{GNN}(G_\text{mol,train})$.
After training, a property prediction for a molecule is classified as reliable or non-reliable by first computing the corresponding molecular fingerprint through the GNN and then evaluating this fingerprint with the SVM.
A molecule is classified as reliable if its molecular fingerprint $\mathbf{h}_\text{FP}$ results in a non-negative value, i.e., $\mli{SVM}_{\text{AD}}(\mathbf{h}_{\text{FP}})~\geq~0$.

Since our GNN is based on ensemble learning (EL), i.e., the predictions of 40 GNNs are averaged to get the final prediction, we train 40 one-class SVMs in total; one SVM for each GNN. 
We then apply a majority vote to determine if a prediction lies within the AD or not (cf. Subsection ``Applicability domain'' in the main text).

\section{Hyperparameters of graph-ML CAMD framework}\label{ESI:HypPar_loop}

\subsection*{Generator and prediction models}
For all three generator models, i.e., JT-VAE~\citep{Jin2018}, MHG-VAE~\citep{Kajino2019}, MolGAN~\citep{DeCao2018}, and the GNN prediction model for fuel ignition quality~\citep{Schweidtmann2020_GNNs}, we use the hyperparameter configuration as provided in the original publications and corresponding code repositories.

\subsection*{Bayesian optimization}
For BO, we follow the optimization procedure of the MHG-VAE study by Kajino~\citep{Kajino2019} and select the default parameters of GPyOpt~\citep{gpyopt2016} using Gaussian Process models with the Matern 5/2 kernel as a surrogate model.
The Gaussian Process surrogate models are initialized with 10 molecules randomly sampled from QM9~\citep{Ruddigkeit2012, Ramakrishnan2014}.
We apply expected improvement for the acquisition function and optimize it with L-BFGS~\citep{Liu.1989}.
In contrast to the MHG study~\citep{Kajino2019}, we use Thompson sampling~\citep{Thompson.1933, Russo.2018} with a batch size of 10 to support exploration.
For the two VAEs, we additionally apply principal component analysis (PCA) from scikit-learn~\citep{scikit-learn} trained on the latent vectors of all QM9 molecules, reducing the dimensionality for the MHG-VAE from 72 to 41 and for the JT-VAE from 56 to 38 which in both cases maintains an explained variance ratio of 99.9\%.

\subsection*{Genetic algorithm}
When using GA for optimization, we apply the default parameters of the package \textit{geneticalgorithm}~\citep{GApip2020}, i.e., mutation and crossover probability of 0.1 and 0.5, respectively, elite ratio of 0.01 and parents portion of 0.3.
However, we reduce the population size from 100 to 50 to increase the number of evolutionary steps per unit computational time and thus exploration.

\subsection*{Applicability domain}
Our implementation of the SVMs for determining the applicability domain (AD) of graph neural networks follows the implementation of Schweidtmann et al.~\citep{Schweidtmann2021_AD}, i.e., we use the class \textit{OneClassSVM} of the Python package scikit-learn~\citep{scikit-learn}. 
We apply the default hyperparameter settings, except for the kernel coefficient $\gamma$ used for radial basis funcion kernel of the SVM and the parameter $\nu$ that determines the maximal fraction of training points classified as outliers and a minimum fraction of support vectors.
We determine $\gamma$ and $\nu$ through a grid search with $\gamma \in \{0.5, 0.1, 0.01, 0.005, 0.001, 0.0005, 0.0001, \text{scale}\}$ and $\nu \in \{0.5, 0.1, 0.05, 0.01\}$.
Note that when using $\gamma =$ scale (default in sklearn), the value for $\gamma$ is automatically selected for each one-class SVM by scikit-learn by multiplying the inverse of the fingerprint dimension with the inverse of the variance in all training molecular fingerprints. 
To identify $\gamma$ and $\nu$, we gradually decrease $\gamma$ while testing different values for $\nu$ until the number of support vectors does not decrease much anymore~\citep{Dreiseitl.2010, Schweidtmann2021_AD}.
We find the default scaling option within skicit-learn for $\gamma$ to work well and values for $\nu$ below 0.05 to not decrease the number of support vectors significantly and thus select $\gamma =  0.00086$ and $\nu = 0.05$ values as final hyperparameters.

\subsection*{Fuel design loop runs}
For all design loop runs, we set the lower and upper bounds of the search space for the optimization to the minimum and the maximum entries of the latent vectors of the generative models, respectively, expanded by 20\% of the difference between maximum and minimum to allow some extrapolation.
The minimum and maximum entries of the latent vectors in the generative models are determined based on the about 50,000 HCO-molecules within the QM9 data set~\citep{Ruddigkeit2012, Ramakrishnan2014} for the two VAEs, JT-VAE and MHG-VAE, and 50,000 samples from a standard normal distribution for MolGAN (cf.~\citep{DeCao2018}).
In addition, we set a maximum time limit of 10 seconds for decoding a latent vector to a molecular graph since we experienced rare cases of very long decoding times with the JT-VAE.
The time limit of 10 seconds corresponds to about 10 times the maximum decoding time of 95~\% of the molecules in QM9 with the JT-VAE, which we found to have the longest decoding times among the generators.

\section{QM9 statistics}\label{ESI:DetRes_opt}

The QM9~\citep{Ruddigkeit2012, Ramakrishnan2014} molecule database that we use for the training of the generative models also includes molecules with high predicted RON and OS (cf. Table~\ref{tab:ResultTable_RONOS_QM9}).
The maximum RON\,$+$\,OS score in QM9 is 135 and the mean score of the top 20 molecules is 129.
When applying the AD, the GNN predictions for about 3,250 out of the total 50,150 hydrocarbons within QM9 are classified as unreliable and are thus omitted, yielding a reduced maximum score of 131 and mean score of 128.

\begin{table}[htbp]
	\centering
	\resizebox{0.9\linewidth}{!}{%
		\begin{tabular}{l|cc}
			\toprule
			predicted RON$+$OS  & QM9   & QM9+AD    \\
			\midrule
			max                 & 135   & 131       \\
			mean top 20         & 129   & 128       \\
			\# unique mol.      & 50150 & 46902     \\
			\# promising mol.   & 63    & 51        \\
			\bottomrule
		\end{tabular}%
	}
	\caption{Promising molecules (both RON\,$>$\,110 and OS\,$>$\,10) in the QM9 data set with and without considering the applicability domain (AD).}
	\label{tab:ResultTable_RONOS_QM9}%
\end{table}

\section{Detailed CAMD results}\label{ESI:DetRes_opt}

\begin{figure*}[b]
	\begin{subfigure}[c]{0.95\textwidth}
		\centering
		\includegraphics[width=\textwidth]{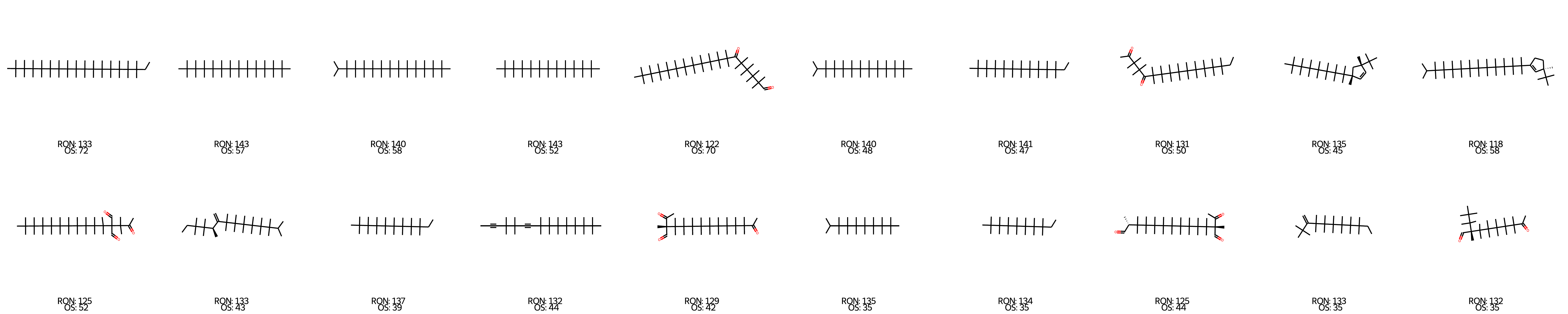}
		\subcaption{JT-VAE, BO}
	\end{subfigure}
	\quad
	\begin{subfigure}[c]{0.95\textwidth}
		\centering
		\includegraphics[width=\textwidth]{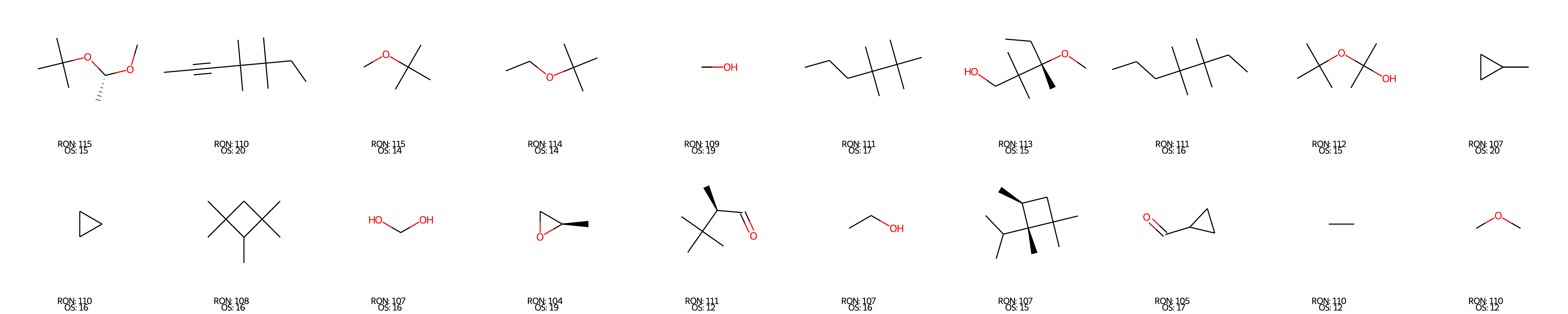}
		\subcaption{JT-VAE, BO+AD}
	\end{subfigure}
	\quad
	\begin{subfigure}[c]{0.95\textwidth}
		\centering
		\includegraphics[width=\textwidth]{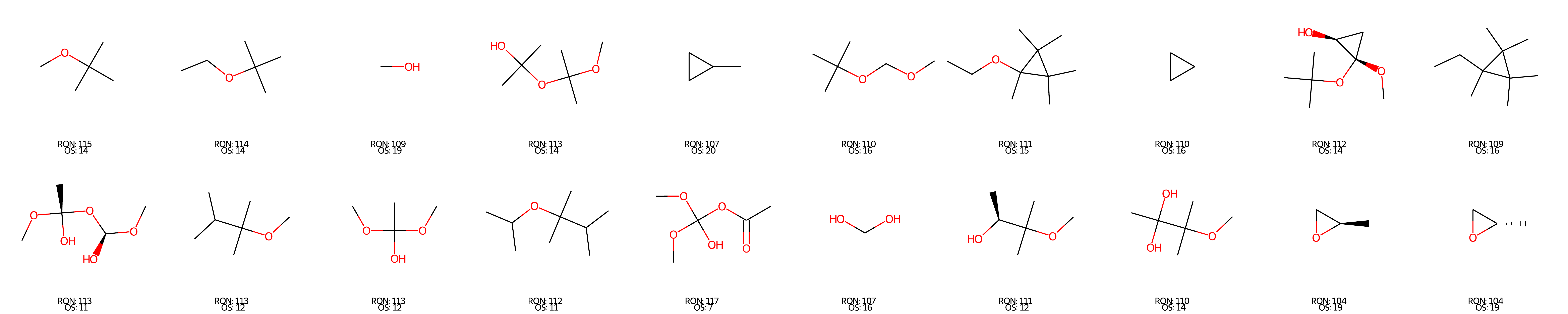}
		\subcaption{JT-VAE, GA}
	\end{subfigure}
	\begin{subfigure}[c]{0.95\textwidth}
		\centering
		\includegraphics[width=\textwidth]{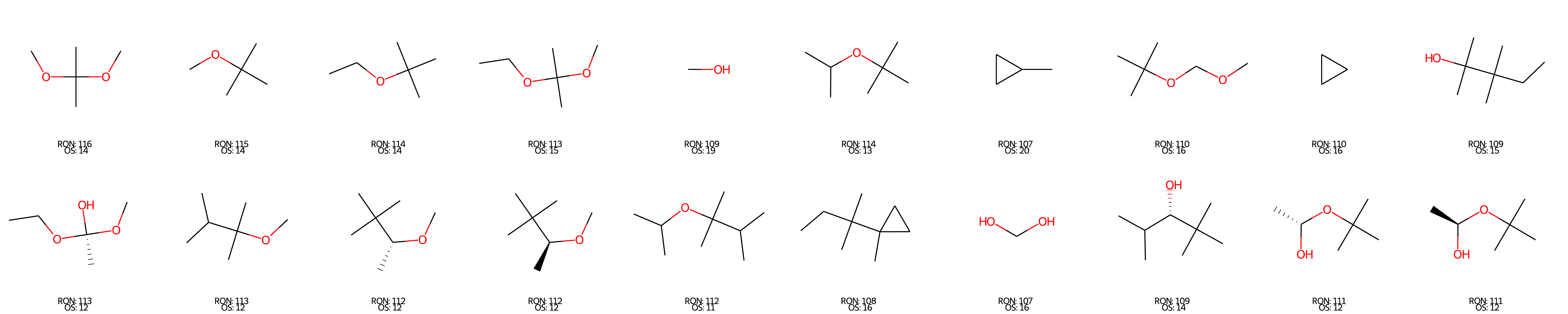}
		\subcaption{JT-VAE, GA+AD}
	\end{subfigure}
	\caption{Top 20 candidates identified in fuel design loop runs with the JT-VAE~\citep{Jin2018} model, with BO and GA, without and with applicability domain, and with stopping criterion SC$_\text{\#molecs}$ (max. 1000 unique molecules, max. 2000 total molecules). RON and OS values are predicted by the graph neural network~\citep{Schweidtmann2020_GNNs}.}
	\label{fig:zESI_Top20_SCi_mols_JTVAE}
\end{figure*}

\begin{figure*}
	\begin{subfigure}[c]{0.95\textwidth}
		\centering
		\includegraphics[width=\textwidth]{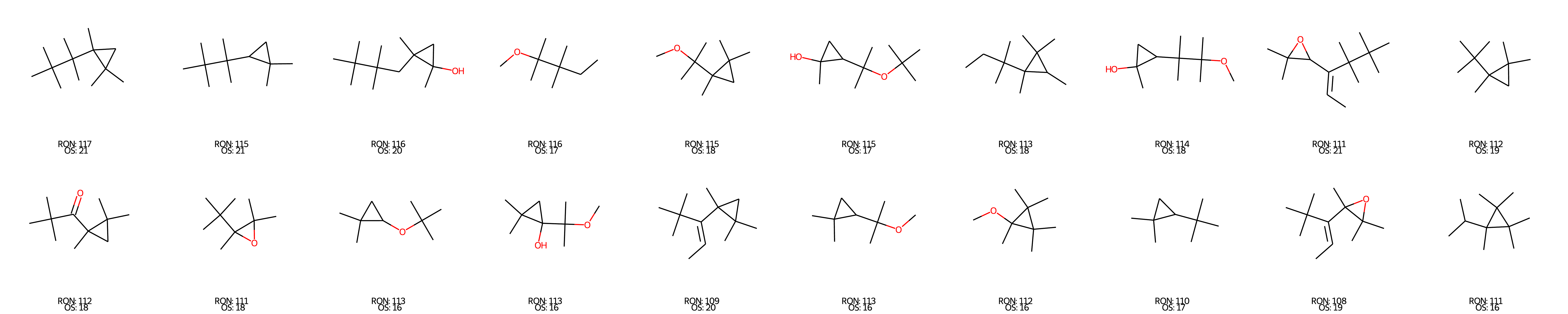}
		\subcaption{MHG-VAE, BO}
	\end{subfigure}
	\quad
	\begin{subfigure}[c]{0.95\textwidth}
		\centering
		\includegraphics[width=\textwidth]{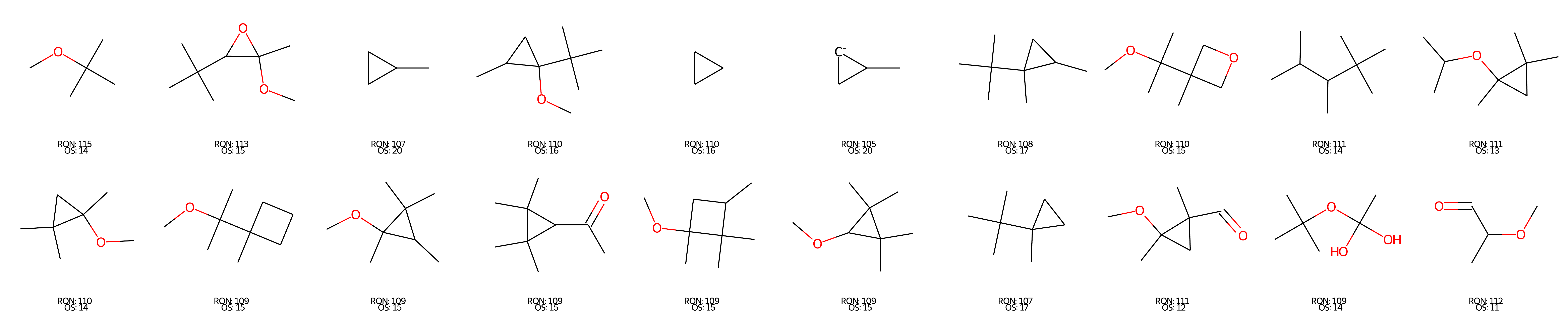}
		\subcaption{MHG-VAE, BO+AD}
	\end{subfigure}
	\quad
	\begin{subfigure}[c]{0.95\textwidth}
		\centering
		\includegraphics[width=\textwidth]{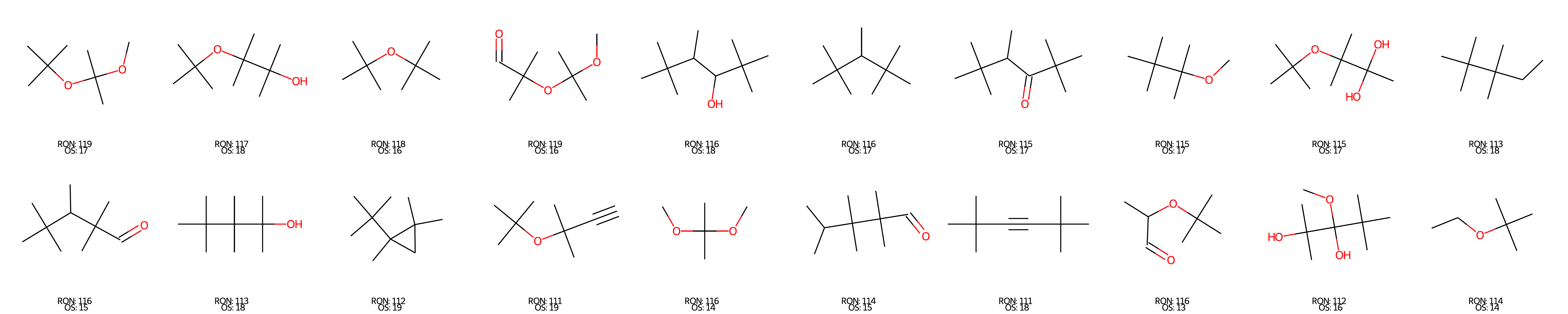}
		\subcaption{MHG-VAE, GA}
	\end{subfigure}
	\quad
	\begin{subfigure}[c]{0.95\textwidth}
		\centering
		\includegraphics[width=\textwidth]{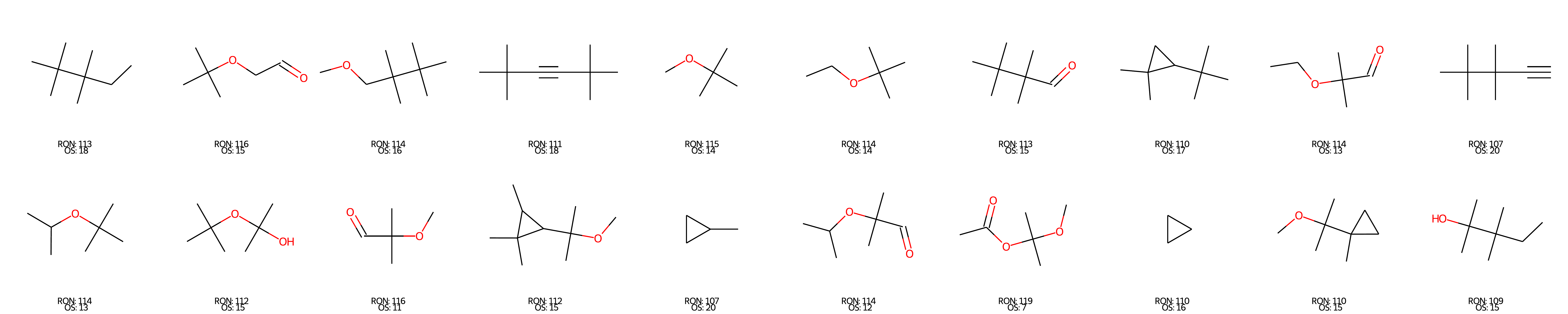}
		\subcaption{MHG-VAE, GA+AD}
	\end{subfigure}
	\caption{Top 20 candidates identified in fuel design loop runs with the MHG-VAE~\citep{Kajino2019} model, with BO and GA, without and with applicability domain, and with stopping criterion SC$_\text{\#molecs}$ (max. 1000 unique molecules, max. 2000 total molecules). RON and OS values are predicted by the graph neural network~\citep{Schweidtmann2020_GNNs}.}
	\label{fig:zESI_Top20_SCi_mols_MHGVAE}
\end{figure*}

\begin{figure*}
	\begin{subfigure}[c]{0.95\textwidth}
		\centering
		\includegraphics[width=\textwidth]{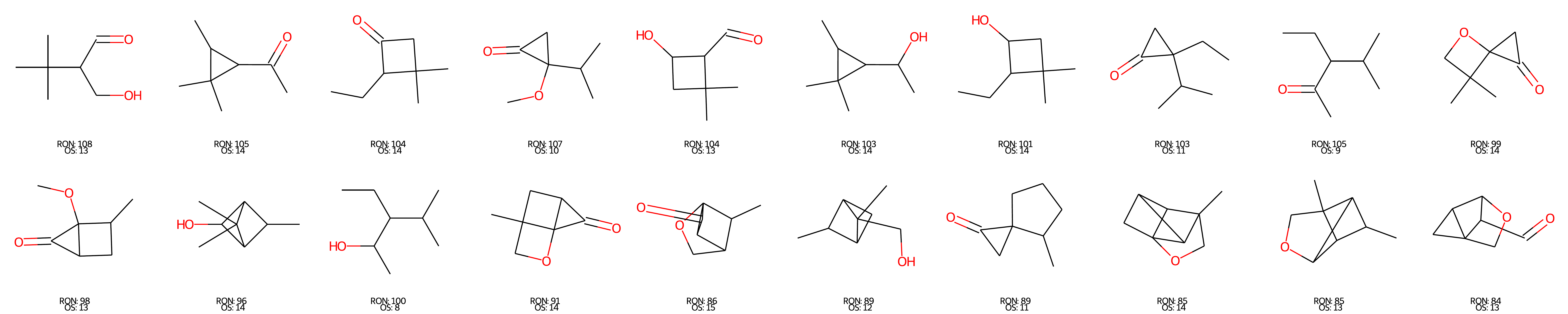}
		\subcaption{MolGAN, BO}
	\end{subfigure}
	\quad
	\begin{subfigure}[c]{0.95\textwidth}
		\centering
		\includegraphics[width=\textwidth]{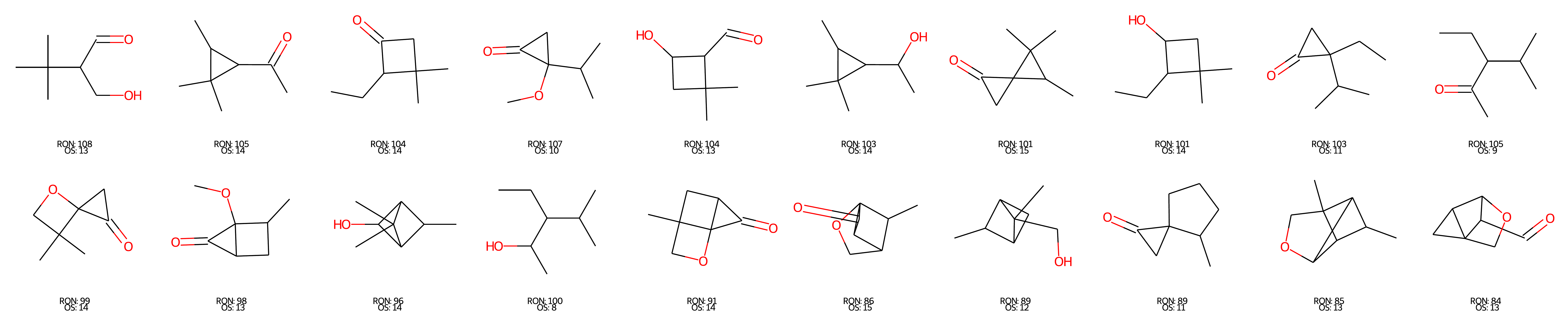}
		\subcaption{MolGAN, BO+AD}
	\end{subfigure}
	\quad
	\begin{subfigure}[c]{0.95\textwidth}
		\centering
		\includegraphics[width=\textwidth]{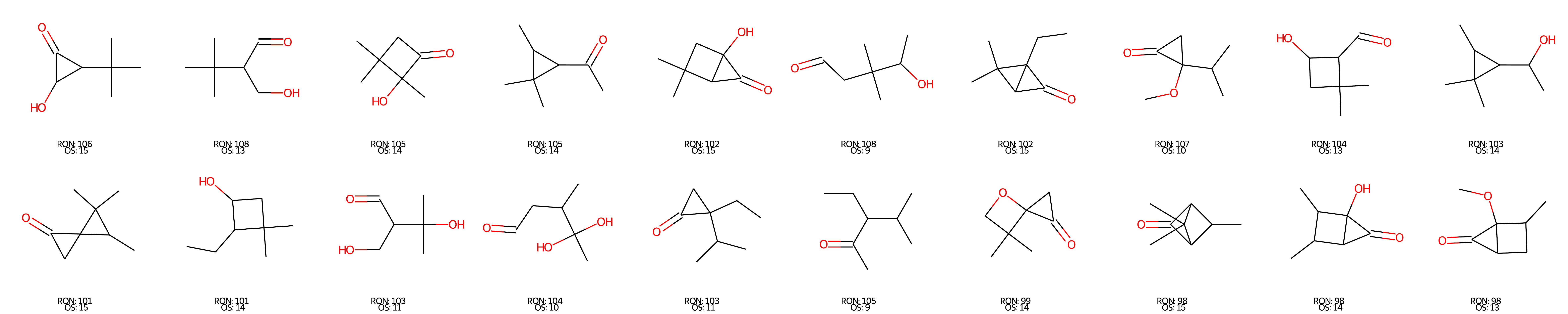}
		\subcaption{MolGAN, GA}
	\end{subfigure}
	\quad
	\begin{subfigure}[c]{0.95\textwidth}
		\centering
		\includegraphics[width=\textwidth]{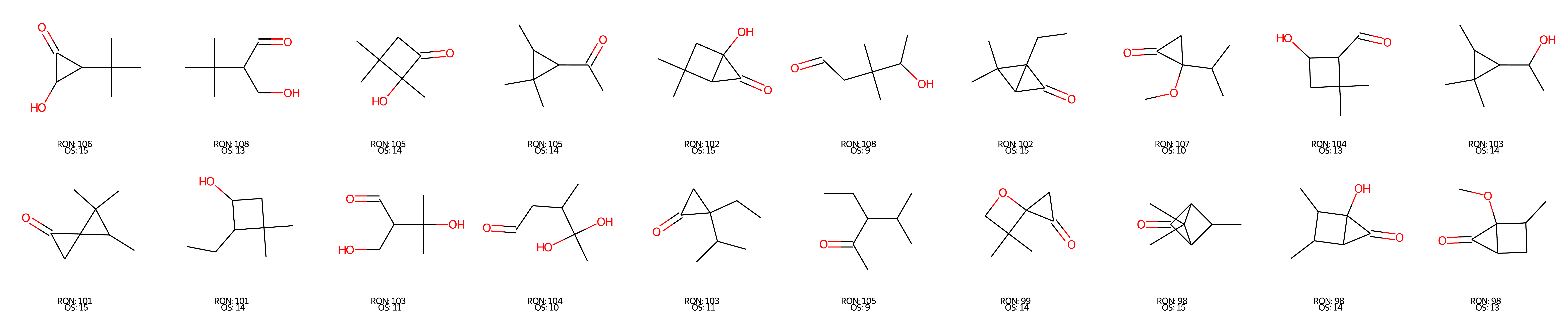}
		\subcaption{MolGAN, GA+AD}
	\end{subfigure}
	\caption{Top 20 candidates identified in fuel design loop runs with the MolGAN~\citep{DeCao2018} model, with BO and GA, without and with applicability domain, and with stopping criterion SC$_\text{\#molecs}$ (max. 1000 unique molecules, max. 2000 total molecules). RON and OS values are predicted by the graph neural network~\citep{Schweidtmann2020_GNNs}.}
	\label{fig:zESI_Top20_SCi_mols_MolGAN}
\end{figure*}

\begin{figure*}
	\begin{subfigure}[c]{0.95\textwidth}
		\centering
		\includegraphics[width=\textwidth]{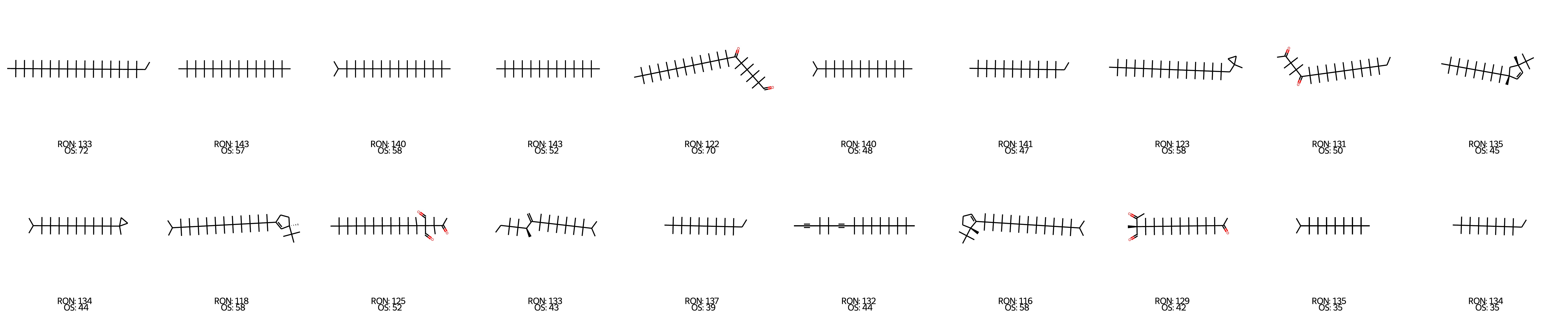}
		\subcaption{JT-VAE, BO}
	\end{subfigure}
	\quad
	\begin{subfigure}[c]{0.95\textwidth}
		\centering
		\includegraphics[width=\textwidth]{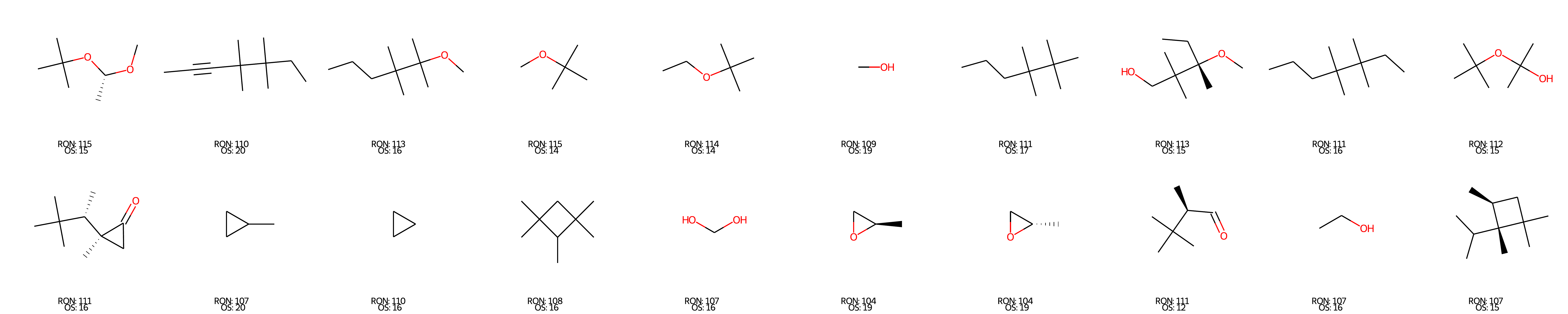}
		\subcaption{JT-VAE, BO+AD}
	\end{subfigure}
	\quad
	\begin{subfigure}[c]{0.95\textwidth}
		\centering
		\includegraphics[width=\textwidth]{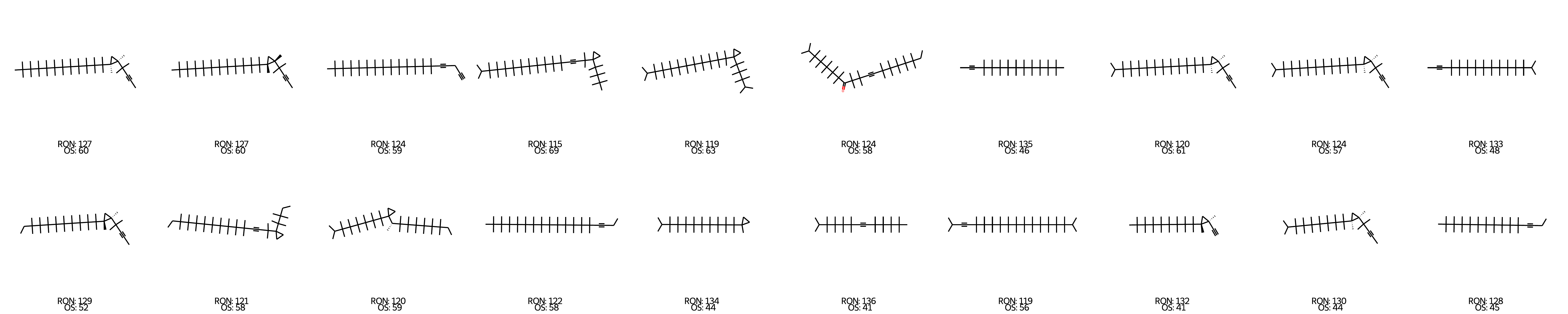}
		\subcaption{JT-VAE, GA}
	\end{subfigure}
	\begin{subfigure}[c]{0.95\textwidth}
		\centering
		\includegraphics[width=\textwidth]{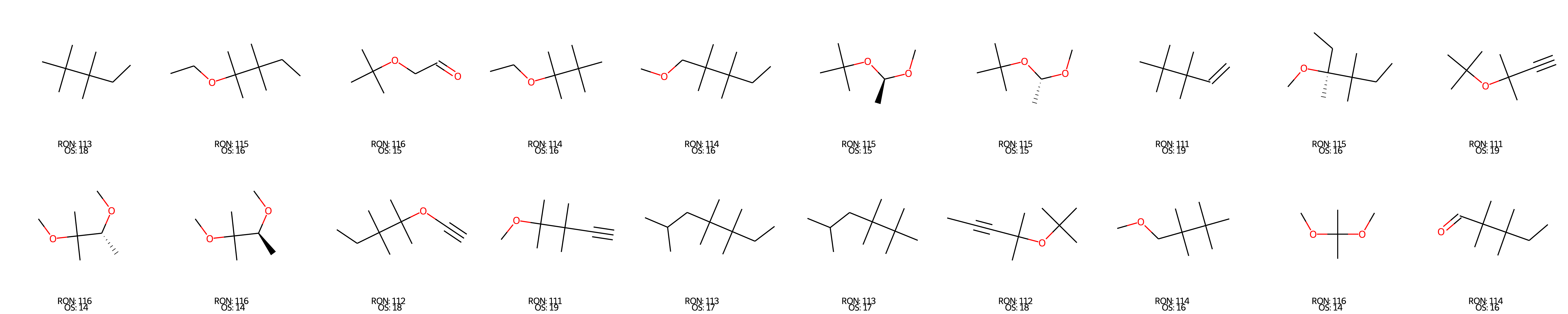}
		\subcaption{JT-VAE, GA+AD}
	\end{subfigure}
	\caption{Top 20 candidates identified in fuel design loop runs with the JT-VAE~\citep{Jin2018} model, with BO and GA, without and with applicability domain, and with stopping criterion SC$_\text{time}$ (12 hours run time). RON and OS values are predicted by the graph neural network~\citep{Schweidtmann2020_GNNs}.}
	\label{fig:zESI_Top20_SCii_mols_JTVAE}
\end{figure*}

\begin{figure*}
	\begin{subfigure}[c]{0.95\textwidth}
		\centering
		\includegraphics[width=\textwidth]{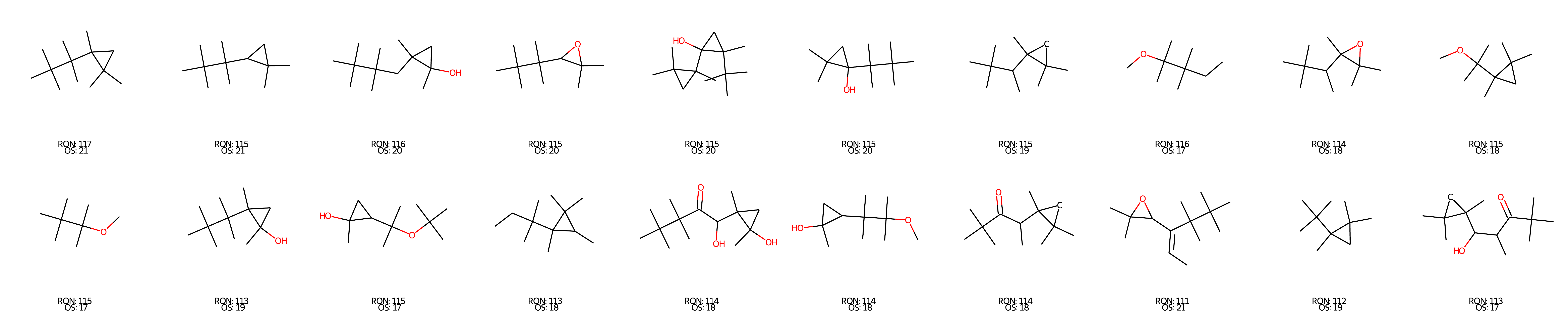}
		\subcaption{MHG-VAE, BO}
	\end{subfigure}
	\quad
	\begin{subfigure}[c]{0.95\textwidth}
		\centering
		\includegraphics[width=\textwidth]{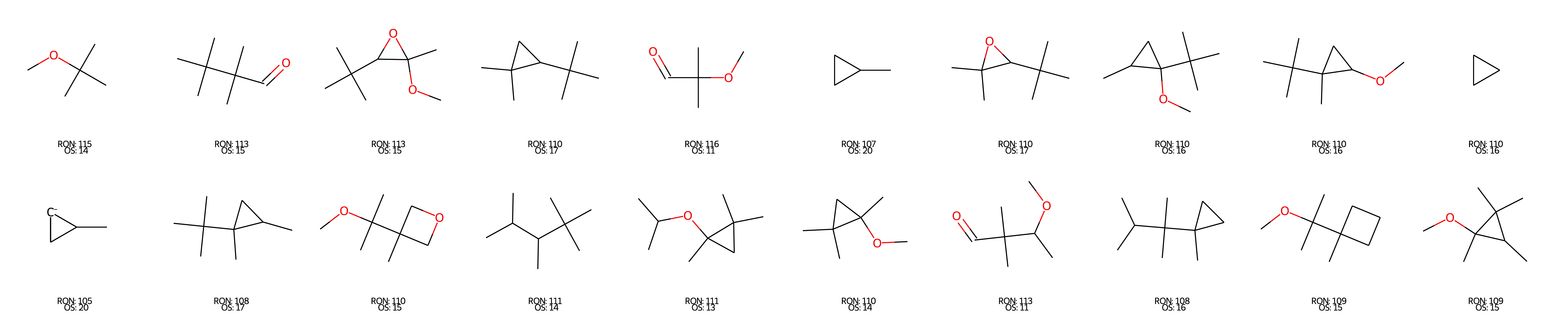}
		\subcaption{MHG-VAE, BO+AD}
	\end{subfigure}
	\quad
	\begin{subfigure}[c]{0.95\textwidth}
		\centering
		\includegraphics[width=\textwidth]{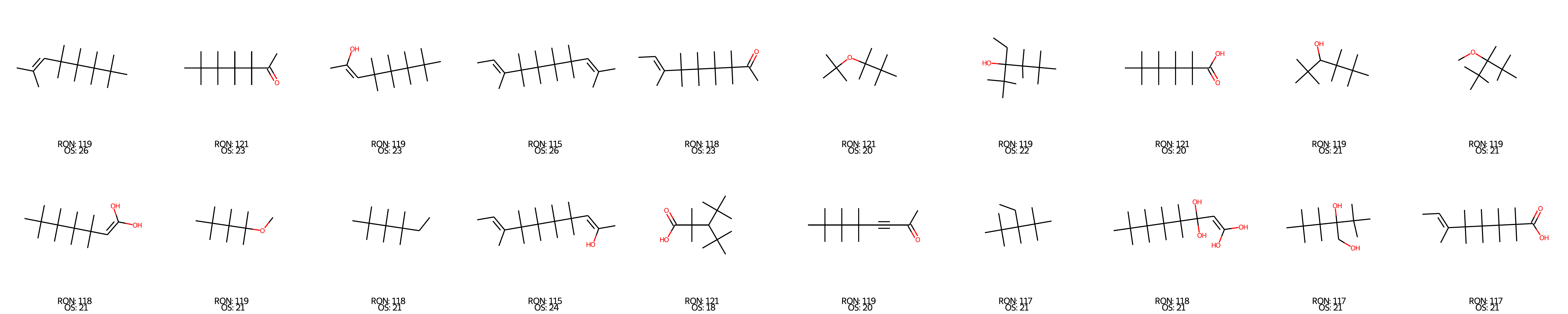}
		\subcaption{MHG-VAE, GA}
	\end{subfigure}
	\quad
	\begin{subfigure}[c]{0.95\textwidth}
		\centering
		\includegraphics[width=\textwidth]{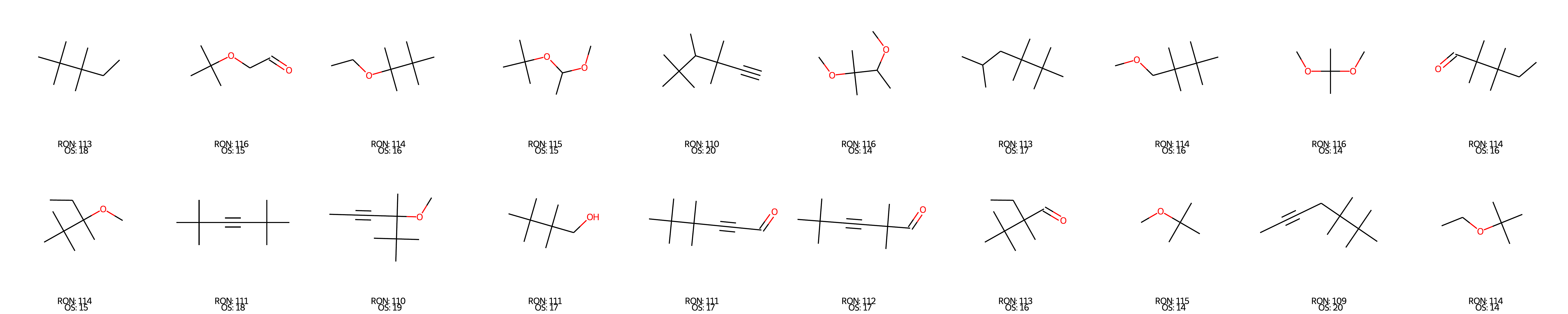}
		\subcaption{MHG-VAE, GA+AD}
	\end{subfigure}
	\caption{Top 20 candidates identified in fuel design loop runs with the MHG-VAE~\citep{Kajino2019} model, with BO and GA, without and with applicability domain, and with stopping criterion SC$_\text{time}$ (12 hours run time). RON and OS values are predicted by the graph neural network~\citep{Schweidtmann2020_GNNs}.}
	\label{fig:zESI_Top20_SCii_mols_MHGVAE}
\end{figure*}

\begin{figure*}
	\begin{subfigure}[c]{0.95\textwidth}
		\centering
		\includegraphics[width=\textwidth]{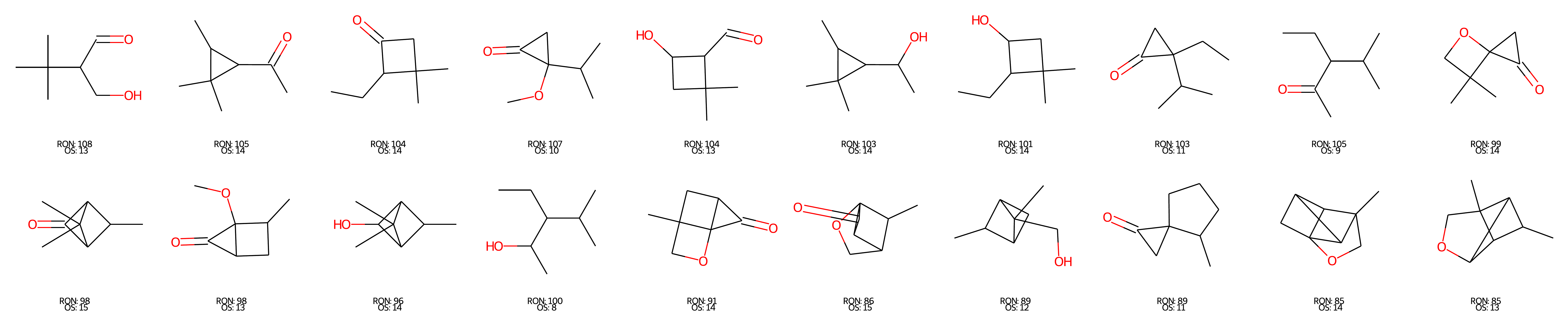}
		\subcaption{MolGAN, BO}
	\end{subfigure}
	\quad
	\begin{subfigure}[c]{0.95\textwidth}
		\centering
		\includegraphics[width=\textwidth]{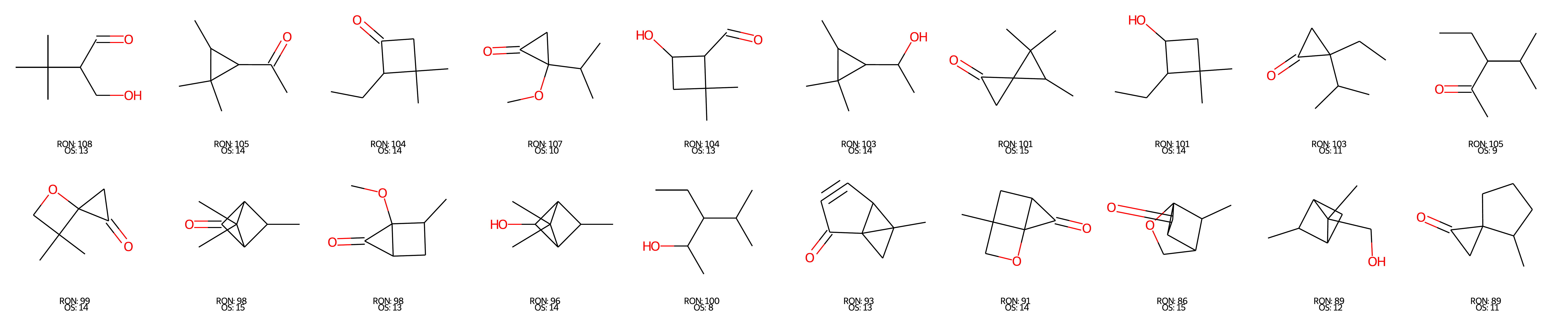}
		\subcaption{MolGAN, BO+AD}
	\end{subfigure}
	\quad
	\begin{subfigure}[c]{0.95\textwidth}
		\centering
		\includegraphics[width=\textwidth]{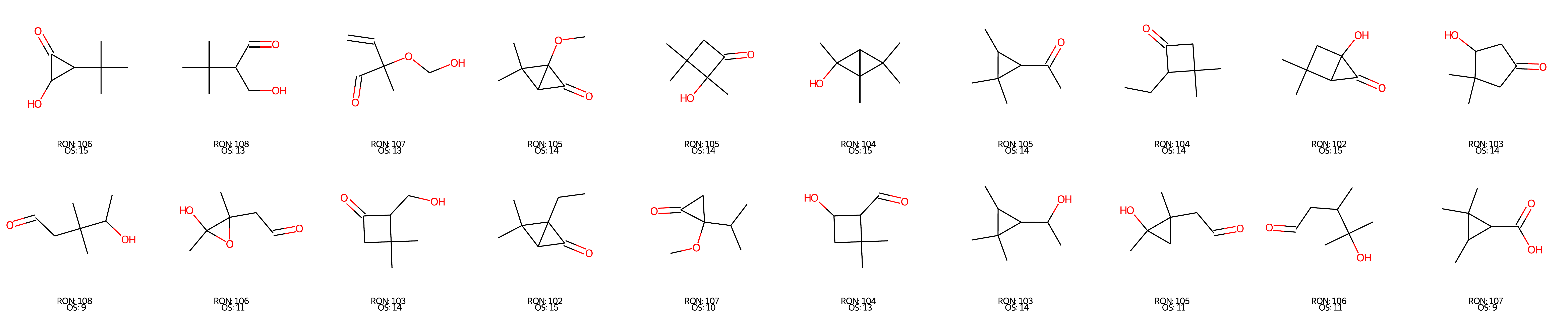}
		\subcaption{MolGAN, GA}
	\end{subfigure}
	\quad
	\begin{subfigure}[c]{0.95\textwidth}
		\centering
		\includegraphics[width=\textwidth]{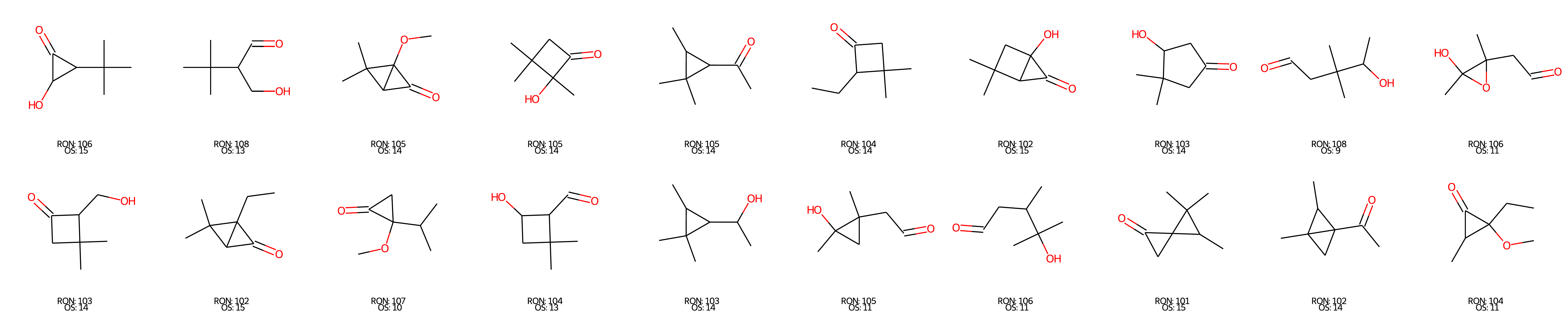}
		\subcaption{MolGAN, GA+AD}
	\end{subfigure}
	\caption{Top 20 candidates identified in fuel design loop runs with the MolGAN~\citep{DeCao2018} model, with BO and GA, without and with applicability domain, and with stopping criterion SC$_\text{time}$ (12 hours run time). RON and OS values are predicted by the graph neural network~\citep{Schweidtmann2020_GNNs}.}
	\label{fig:zESI_Top20_SCii_mols_MolGAN}
\end{figure*}

\begin{figure*}
	\begin{subfigure}[c]{0.95\textwidth}
		\centering
		\includegraphics[width=\textwidth]{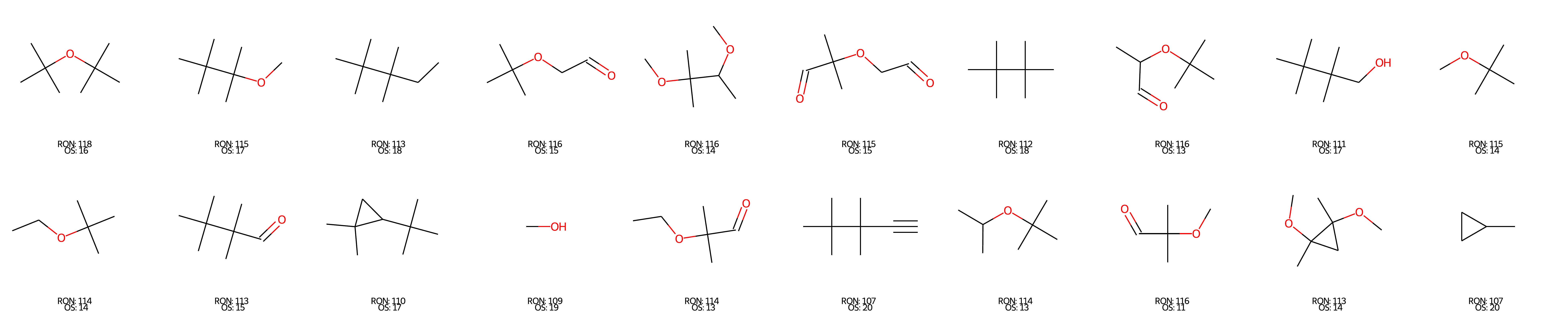}
		\subcaption{QM9}
	\end{subfigure}
	\quad
	\begin{subfigure}[c]{0.95\textwidth}
		\centering
		\includegraphics[width=\textwidth]{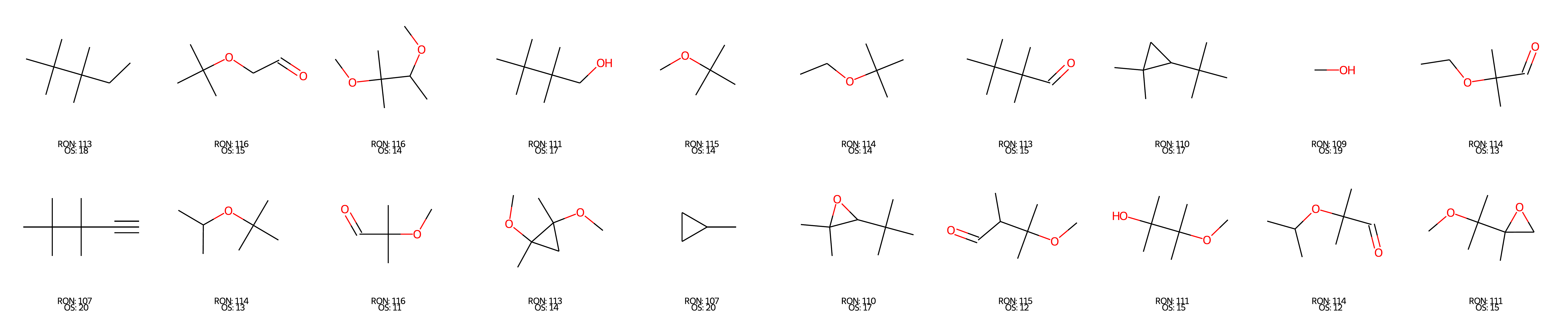}
		\subcaption{QM9+AD}
	\end{subfigure}
	\caption{Top 20 molecular candidates with regard to predicted \ronos{} within the QM9 data set~\citep{Ruddigkeit2012, Ramakrishnan2014} without and with applicability domain. RON and OS values are predicted by the graph neural network~\citep{Schweidtmann2020_GNNs}.}
	\label{fig:TopMols_QM9}
\end{figure*}

Figures~\ref{fig:zESI_Top20_SCi_mols_JTVAE}-\ref{fig:zESI_Top20_SCi_mols_MolGAN} show the top 20 molecules with regard to predicted \ronos{} that are identified in the fuel design loop runs with the three generator models (JT-VAE~\citep{Jin2018}, MHG-VAE~\citep{Kajino2019}, MolGAN~\citep{DeCao2018}) and the two optimization approaches (BO and GA) with and without AD and with stopping criterion SC$_\text{\#molecs}$ (max. 1000 unique molecules, max. 2000 total molecules).
Figures~\ref{fig:zESI_Top20_SCii_mols_JTVAE}-\ref{fig:zESI_Top20_SCii_mols_MolGAN} show the top 20 molecules with regard to predicted \ronos{} that are identified in the fuel design loop runs with the respective generator and optimization approaches with and without AD and with stopping criterion SC$_\text{time}$ (12 hours run time).
Figure~\ref{fig:TopMols_QM9} illustrates the top 20 molecules with regard to predicted \ronos{} in the QM9 dataset~\citep{Ruddigkeit2012, Ramakrishnan2014}, for both with and without AD.

\clearpage

\section{Rapid compression machine screening}
\begin{figure}
	\centering
	\includegraphics[width=0.5\textwidth, trim={1cm 17cm 1cm 1cm},clip]{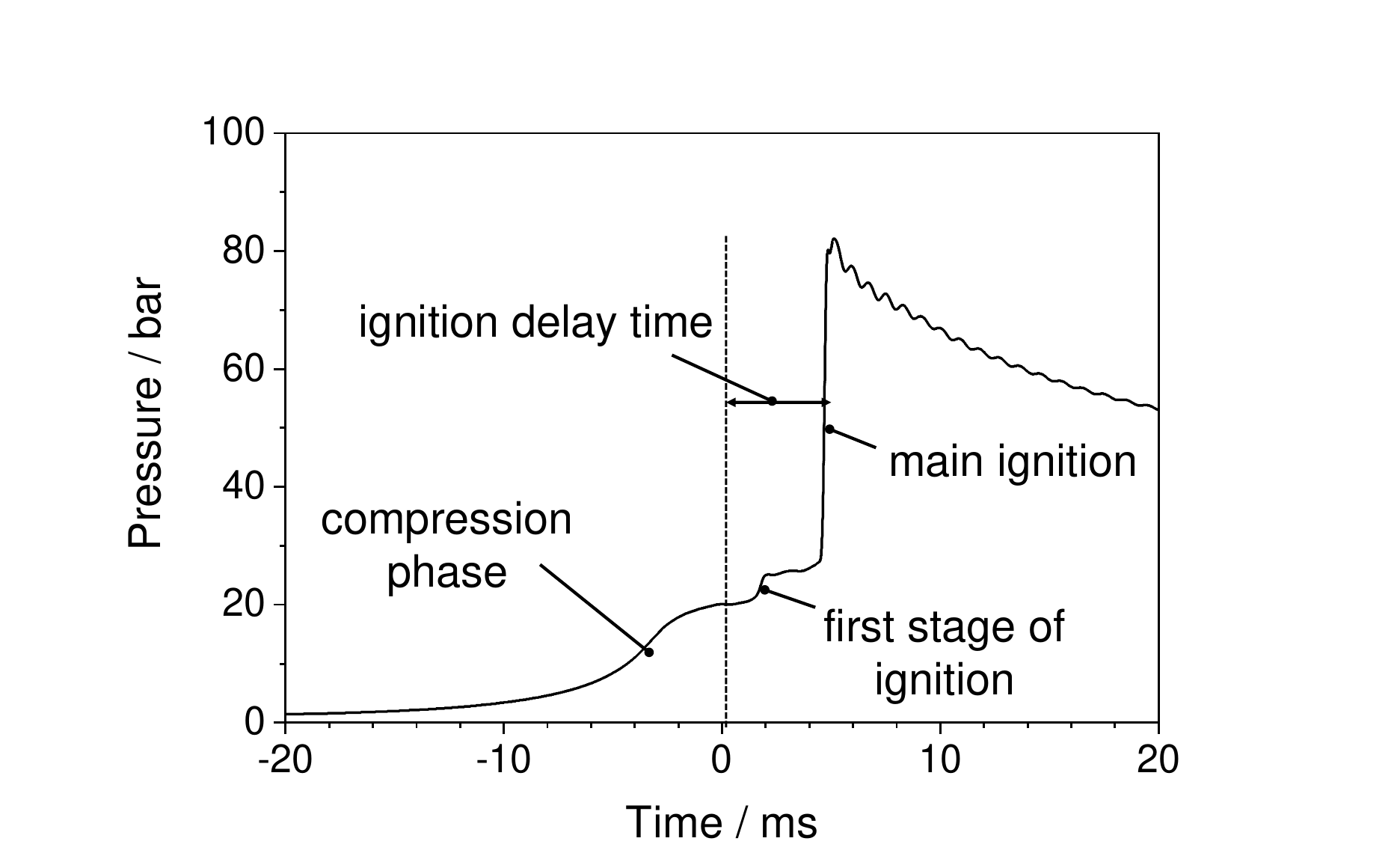}
	\caption{Exemplary pressure trace of a rapid compression machine experiment with a two-stage ignition. Time 0 corresponds to the end of compression.}
	\label{fig:RCM_exp}
\end{figure}

Ignition delay times were measured in our rapid compression machine (RCM) at RWTH Aachen University.
The RCM has been presented in detail by Lee et al.~\citep{LEE2012} and Ramalingam et al.~\citep{RAMALINGAM2017}.
Briefly, the RCM has a stainless steel reaction chamber with an inner diameter of 50 mm and a total stroke length of 250 mm. 
Both the manifold system with the two mixture storage tanks and the reaction chamber were heated to 75 °C during the experimental series to avoid local condensation issues. 
A Kistler pressure transducer (6125C-U20) was used to determine the end-of-compression (EOC) pressure and to detect the ignition event, which was defined by the maximum pressure gradient with respect to time.
The time interval between these events was defined as the ignition delay time, cf. Figure~\ref{fig:RCM_exp}. 
In order to change the compression ratio and thus EOC temperature, which was calculated under the assumption of an isentropic compression inside the core gas of the reactor, a moveable endwall was used. 
The measurement uncertainties for the EOC conditions have been estimated to be ± 5 K for the EOC temperature and ± 0.15 bar for the EOC pressure, respectively~\citep{RAMALINGAM2017}. 
The EOC pressures of 20 and 40 bar were set by adjusting the initial pressure in the reaction chamber accordingly. 
A dilution of 3.762 was achieved by diluting a stoichiometric mixture of 2,2-dimethoxypropane (Alfa Aesar, 99.7 \% purity) and oxygen (grade 5.0) with argon (grade 5.0) ensuring a constant mixture composition for the measurable temperature regime. 
The expected scatter of ± 20 \%~\citep{RAMALINGAM2017} is indicated by the error bars in the illustration of the experimental results.

  \clearpage
  \newpage

  \bibliographystyle{apalike}
  \renewcommand{\refname}{Bibliography}  
  \bibliography{literature.bib}